\documentclass[10pt,twocolumn,letterpaper]{article}

 \usepackage{cvpr}              
\usepackage[accsupp]{axessibility}

 \usepackage{multirow}
 \usepackage{multicol}
 \usepackage{ragged2e}
 \usepackage{bm}
 \usepackage{wrapfig}
 \usepackage{enumitem}
 \usepackage{listings}

\definecolor{cvprblue}{rgb}{0.21,0.49,0.74}
\usepackage[pagebackref,breaklinks,colorlinks,allcolors=cvprblue]{hyperref}

\def\paperID{} 
\def\confName{CVPR}
\def\confYear{2026}

\title{GeoSANE: Learning Geospatial Representations from Models, Not Data}

\author{
Joëlle Hanna$^{1}$ \quad Damian Falk$^{1}$ \quad Stella X. Yu$^{2}$ \quad Damian Borth$^{1, 3}$ \\
$^{1}$ University of St.Gallen \space\space\space $^{2}$ 
University of Michigan and UC Berkeley \space\space\space $^{3}$ ESA $\Phi$-Lab\\
}

\begin{document}
\maketitle
\begin{abstract}
Recent advances in remote sensing have led to an increase in the number of available foundation models;  each trained on different modalities, datasets, and objectives, yet capturing only part of the vast geospatial knowledge landscape. While these models show strong results within their respective domains, their capabilities remain complementary rather than unified. Therefore, instead of choosing one model over another, we aim to combine their strengths into a single shared representation.
We introduce GeoSANE, a geospatial model foundry that learns a unified neural representation from the weights of existing foundation models and task-specific models, able to generate novel neural networks weights on-demand. Given a target architecture, GeoSANE generates weights ready for finetuning for classification, segmentation, and detection tasks across multiple modalities.
Models generated by GeoSANE consistently outperform their counterparts trained from scratch, match or surpass state-of-the-art remote sensing foundation models, and outperform models obtained through pruning or knowledge distillation when generating lightweight networks. Evaluations across ten diverse datasets and on GEO-Bench confirm its strong generalization capabilities.
By shifting from pre-training to weight generation, GeoSANE introduces a new framework for unifying and transferring geospatial knowledge across models and tasks. Code is available at \href{https://hsg-aiml.github.io/GeoSANE/}{hsg-aiml.github.io/GeoSANE/}. 

\begin{figure}[t]
    \centering
    \includegraphics[scale=0.255]{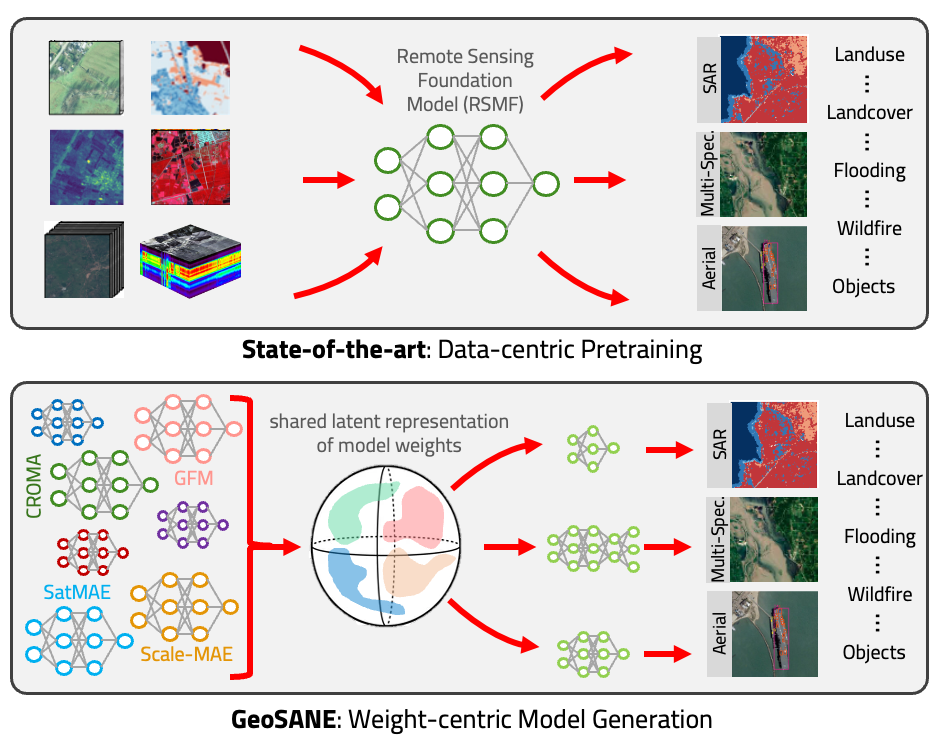}
    \vspace{-0.5cm}
    \caption{Instead of pretraining geospatial (foundation) models from satellite data (top), GeoSANE proposes to leverage publicly available geospatial (foundation) models to train a shared latent representation encapsulating their knowledge to generate model weights tailored for specific downstream task (bottom).}
    \label{fig:front-pager}
    \vspace{-0.25cm}
\end{figure}

\end{abstract}

%
%
%
%
%
%
\vspace{-1em}
\section{Introduction}
\label{sec:intro}
%
%
Foundation models have transformed computer vision and remote sensing by providing strong, task-agnostic representations capable of adapting towards a diverse set of downstream tasks. In remote sensing, foundation models~\cite{Cong2022SatMAEPT, Reed2022ScaleMAEAS, Prithvi-100M-preprint, Szwarcman2024PrithviEO20AV, Xiong2024NeuralPM, Mapex2026Hanna, Guo2023SkySenseAM, Jakubik2025TerraMindLG, Danish2025TerraFMAS, Fuller2023CROMARS, Astruc2024AnySatOE}  have demonstrated that large-scale pretraining on multispectral and multimodal satellite data yields transferable representations for classification, segmentation, and detection. 
%
%
However, despite their success, the landscape of remote sensing foundation models (RSFMs) remains fragmented: each model specializes in a subset of sensors, spatial resolutions, or objectives, with many models being complementary and some being more comprehensive than others. Recent surveys~\cite{Xiao2024FoundationMF, Lu2024VisionFM} count more than 70 RSFMs, with more RSFM joining this list. 
As a result, users must repeatedly decide which RSFM to select or retrain for a new task, despite the fact that the union of all these models probably encapsulates a broader range of geospatial knowledge than any single RSFM alone.

%
%
In this work, we propose to take a fundamentally different perspective. Instead of learning yet another foundation model from remote sensing data, we propose to learn from existing models themselves — directly from their parameters in weight space. Inspired by recent advances in \textit{weight space learning}~\cite{Schrholt2021HyperRepresentationsSR, Schrholt2022HyperRepresentationsAG, Peebles2022LearningTL, Schrholt2024TowardsSA, Wang2024NeuralND, Soro2024DiffusionbasedNN}, we treat the weights of trained neural networks as input modality to learn a single shared latent representation of a given population of neural network models~\cite{schurholt2024neural, han2026survey}. This approach would allow us to combine the knowledge encoded in multiple pretrained neural networks such that one could efficiently generate new model weights, being more suitable for task-specific fine-tuning but without the cost of large-scale pretraining.
\newpage 

%
%
We introduce GeoSANE, a \underline{s}equential \underline{a}utoencoder for \underline{n}eural \underline{e}mbeddings acting as a geospatial model foundry that learns a shared latent representation across diverse RSFMs and task-specific remote sensing models. GeoSANE leverages a transformer-based encoder-decoder in weight space to embed these heterogeneous neural networks spanning different architectures, sensing modalities, and objectives, into a shared latent manifold. From this manifold, GeoSANE can sample new weights to generate entire models for a target architectures and a given remote sensing downstream task. 
In doing so, GeoSANE shifts the paradigm from \textit{data-centric pretraining} to \textit{weight-centric  generation} of remote sensing models (Fig. \ref{fig:front-pager})

%
%
The proposed approach is motivated by three key points:
(i) The availability of open-source geospatial models on platforms such as Hugging Face provides a rich collection of pretrained models capturing a vast amount of domain-specific knowledge.
(ii) Operating directly in weight space decouples knowledge transfer from data availability, enabling efficient adaptation across sensors, modalities, and tasks. 
(iii) Pretrained models constrain their task-specific models to be of similar architecture and size demanding an additional distillation step to build lightweight models.
GeoSANE aims to address all the points above, as it collects a heterogeneous population of existing geospatial models, tokenizes their parameters, and learns a shared latent embedding across them. Given a target architecture (being small or large), GeoSANE samples from this latent space to produce functional models whose weights encode the aggregated knowledge of the full population. 

We evaluate GeoSANE on 10 remote sensing datasets spanning optical, multispectral, and radar modalities, and across classification, segmentation, and detection tasks. Generated models consistently outperform training from scratch, match or exceed leading foundation models, and outperform pruning and distillation baselines when used to create lightweight neural networks. These results demonstrate strong generalization and the feasibility of model generation in weight space for geospatial domains.
In summary, our contributions are threefold:
\begin{itemize}
    \item A new paradigm for remote sensing pretraining. We propose to learn from existing geospatial models in weight space rather than from remote sensing data.
    \item GeoSANE, a scalable encoder-decoder approach that trains from heterogeneous models and enables on-demand weight generation for arbitrary architectures, modalities, and tasks.
    \item Comprehensive empirical validation across multiple tasks, sensors, and datasets, establishing weight space learning as a viable alternative to regular pretraining.\looseness=-1
\end{itemize}

Together, these results indicate that learning from existing models in weight space provides a novel path to harness the rich landscape of geospatial foundation models.

\begin{figure*}[t]
    \centering
\includegraphics[width=1.0\linewidth]{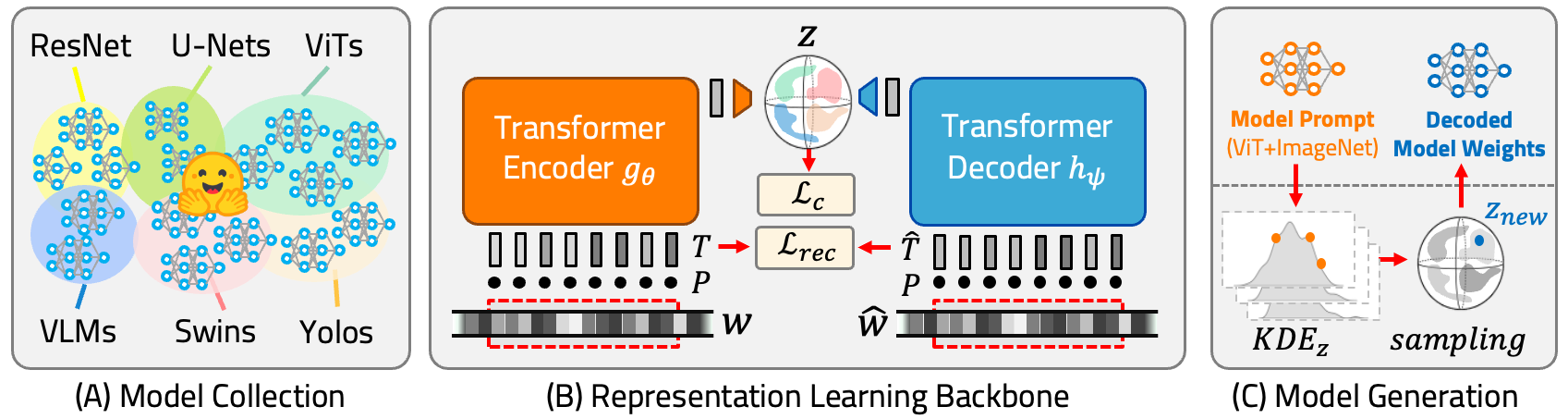}
    \caption{Overview of our approach. (A) A heterogeneous collection of models, including ViTs, Swins, ResNets, UNets, and vision-language models, is gathered from HuggingFace.
(B) A weight-space autoencoder is trained to reconstruct and embed these models into a shared latent representation.
(C) From this latent space, GeoSANE can generate new models on demand for specific downstream tasks such as flood segmentation, object detection, or land-cover classification.}
    \label{fig:pipeline}
    \vspace{-0.3cm}
\end{figure*}

\section{Related Work}
\label{sec:relatedwork}

\vspace{-0.1cm}
\paragraph{Remote Sensing Foundation Models.}
\label{sec:relatedwork:rsfms}
Recent progress in remote sensing foundation models (RSFMs) enabled the reuse of pretrained models for many different satellite image analysis tasks~\cite{scheibenreif2022self, he2025research, hanna2025know}. Early models such as SeCo~\cite{Maas2021SeasonalCU}, SSL4EO-S12~\cite{Wang2022SSL4EOS12AL}, self-supervised ViT \cite{scheibenreif2022self} and SatMAE~\cite{Cong2022SatMAEPT} demonstrated the benefits of large-scale pretraining on Sentinel imagery~\cite{torres2012gmes, drusch2012sentinel} for downstream classification and segmentation. Later works have greatly expanded model capacity and data diversity. ScaleMAE~\cite{Reed2022ScaleMAEAS} introduced scale-aware pretraining for multi-resolution data, while Prithvi-EO~\cite{Prithvi-100M-preprint} and its successor Prithvi-EO-2.0~\cite{Szwarcman2024PrithviEO20AV} leveraged multimodal Landsat-Sentinel fusion to build general-purpose geospatial transformers.
More recent foundation models have grown larger and more versatile, covering more sensors, data types, and tasks. CROMA~\cite{Fuller2023CROMARS} proposed cross-modal alignment between optical and radar inputs to improve multimodal understanding. DOFA~\cite{Xiong2024NeuralPM} introduced domain-oriented pretraining for aerial imagery, emphasizing generalization across data sources. RingMo~\cite{Sun2023RingMoAR} explored large-scale multimodal transformers trained on diverse global datasets, while AnySat~\cite{Astruc2024AnySatOE} aimed for universality across sensors and resolutions. Building on these trends, TerraFM~\cite{Danish2025TerraFMAS} scaled model and data size to a continental level with multisensor pretraining, TerraMind~\cite{Jakubik2025TerraMindLG} extended this direction toward generative, multimodal foundation models, and MAPEX~\cite{Mapex2026Hanna} looked into multimodal Mixture-of-Expert (MoE) setups and expert pruning. 
\\
\indent In contrast, GeoSANE moves beyond traditional satellite imagery based pretraining by learning directly from existing remote sensing model weights to combine their encoded knowledge in an shared latent representation.

%
%
\vspace{-0.3cm}
\paragraph{Model Weight Generation.}
\label{sec:relatedwork:weights}
Early work on parameter generation used hypernetworks~\cite{Ha2016HyperNetworks}, where a separate network predicts the weights of a target model. More recent work instead treats trained weights as a data modality, learning directly from weight space. Hyper-Representations~\cite{Schrholt2021HyperRepresentationsSR} showed that one can learn a lower-dimensional manifold from a population of neural network models and that this shared latent representation can be exploited to generate functional models~\cite{Schrholt2022HyperRepresentationsAG, Schrholt2024TowardsSA}. Recent work further demonstrated that such representations can be learned from heterogeneous models hosted on hubs~\cite{Falk2025LearningMR}, removing the need for curated model zoos~\cite{schurholt2022model, falk2025vitzoo, schurholt2025phasetransitions}.
%
%
More recent diffusion-based approaches, including G.pt~\cite{Peebles2022LearningTL}, RPG~\cite{wang2025scaling} and D2NWG~\cite{Soro2024DiffusionbasedNN}, model the distribution of trained weights to generate parameters conditioned on dataset or architecture, demonstrating the feasibility of sampling performant networks directly in weight space. 
\\
\indent Despite these advances, learning from weights mostly remains limited to either homogeneous model populations~\cite{Peebles2022LearningTL, Schrholt2022HyperRepresentationsAG, Schrholt2024TowardsSA, wang2025scaling, Soro2024DiffusionbasedNN} or traditional computer vision models~\cite{Falk2025LearningMR}. This work addresses, for the first time, learning from weights of fundamentally different architectures processing non-RGB based input modalities and goes beyond image classification to pixel-wise segmentation and object detection.

\vspace{-0.3cm}
\paragraph{Model Merging.}
\label{sec:relatedwork:merging}
Model merging enables the combination of multiple models into a single one. Early work focused on weight-space ensembling, where models fine-tuned from a shared initialization are combined by averaging or interpolating their parameters, as demonstrated in Model Soups~\cite{Wortsman2022ModelSA} and WiSE-FT~\cite{Wortsman2021RobustFO}. However, when the merged models have been fine-tuned on divergent tasks, simple averaging often leads to performance degradation due to conflicting parameter updates. To address these limitations, recent methods like TIES-Merging~\cite{Yadav2023TIESMergingRI} prune insignificant weight changes and align important updates before averaging, while DARE~\cite{Yu2023LanguageMA} zeroes small deltas and amplifies larger ones to reduce interference. Nonetheless, most of these techniques assume that all models share the same architecture and initialization.

In contrast, GeoSANE is designed to encapsulate the knowledge of a large collection of remote sensing models, regardless of their architecture or initialization, overcoming key limitations of prior model merging methods.

\section{Method}
\label{sec:methods}
GeoSANE works in three stages (Fig. \ref{fig:pipeline}). First, we collect a heterogeneous collection of remote sensing models that cover different architectures, tasks, and sensing modalities. Next, we train a weight-space autoencoder to embed these models into a shared latent representation. Finally, given a user defined prompt model, we use the learned latent space to generate new weights for the same architecture, producing models that are ready for fine-tuning on downstream tasks.
We describe each stage in the sections that follow.

\subsection{Remote Sensing Model Collection}
\label{sec:methods:datacollection}
\paragraph{Model Retrieval.}
To train GeoSANE, we need a diverse set of remote sensing models that capture knowledge across multiple sensing modalities, tasks, and architectures. To create such a dataset, we leverage the HuggingFace Hub, which hosts an increasing number of open-source open-weights geospatial models. We query the hub using a broad range of keywords and tags describing both sensing modalities (e.g., Sentinel-1, Sentinel-2, SAR, multispectral) and tasks (e.g., land cover mapping, land cover segmentation, flood detection, disaster response). The complete list of keywords and tags used for model retrieval is provided in the supplementary material.
GeoSANE is designed to automatically load and process a wide variety of architectures, including Transformer-based backbones (ViT, Swin, etc.), CNNs (ResNet, UNet, MobileNet, etc.), multimodal radar-optical models, YOLO-style detectors, task-specific models for floods and wildfires, and vision-language models. We exclude corrupted checkpoints, highly undocumented repositories and models requiring unsafe remote code execution. For models with custom or non-standard implementations such as those from TorchGeo~\cite{Stewart2021TorchGeoDL} or FLAIR~\cite{Garioud2023FLAIRAC}, we implement custom model loaders, which we will make publicly available together with the final model collection.


\paragraph{Final Model Collection.}
After filtering, the final collection contains 103 (foundation) remote sensing models, representing approximately 38 billion parameters. Although the number of individual models appears to be smaller than usually found machine learning datasets, the number of model parameters per model is large ranging between hundreds of millions of parameters to billions of parameters. 



\begin{figure}[t]
    \centering
    \includegraphics[width=1.0\linewidth]{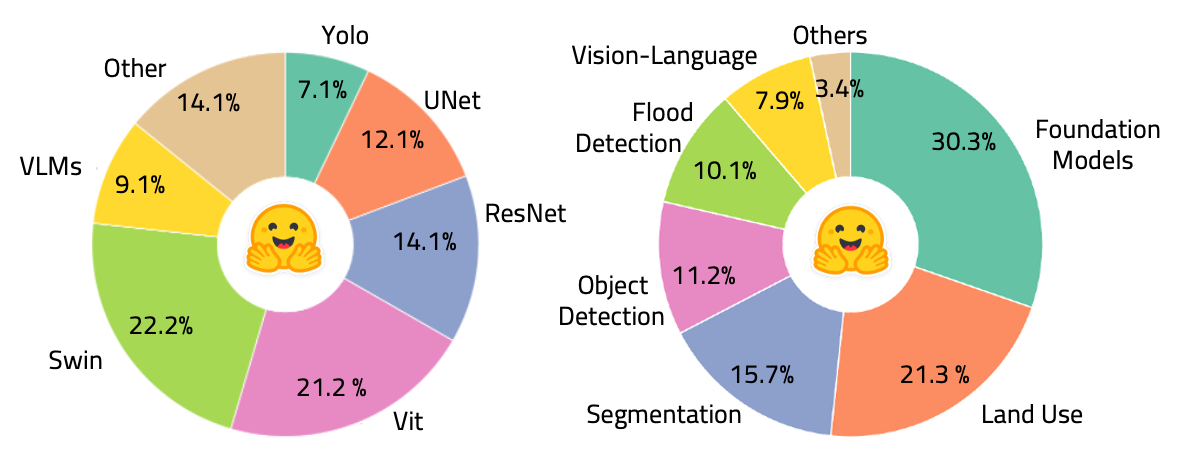}
    \caption{Our model collection retrieved from Hugging Face is diverse as seen in the distribution of model categories in our dataset.}
    \label{fig:dataset_pie}
    \vspace{-1em}
\end{figure}

As a result, even a moderate number of models yields a fair amount of tokens and parameters overall (see Section \ref{sec:experiments:pretraining_implementation}), which is sufficient to learn a strong shared representation.
As illustrated in Figure \ref{fig:dataset_pie}, our dataset collection covers a wide range of model categories, including both foundation and task-specific models, providing a comprehensive and representative view of remote sensing models landscape.
\noindent
This model collection serves as the training dataset for GeoSANE, i.e., as input to learn a shared latent representation of neural network weights.

\subsection{Learning the Shared Latent Representation}
\label{sec:methods:geosane}
While there are various weight space learning methods for model weights generation (see Section \ref{sec:relatedwork:weights}), GeoSANE follows the encoder-decoder setup of~\cite{Schrholt2024TowardsSA} due to its capability to scale to larger model sizes. The core idea of~\cite{Schrholt2024TowardsSA} is to tokenize model weights and express an entire model as a sequence of token vectors. Using such a configuration allows the encoder-decoder backbone to learn representations on chunks of the sequences, and therefore enables generating model sequences of different lengths, underlying architectures, and sizes. 


\vspace{-1em}
\paragraph{Tokenization of Model Weights.}
To that end, the weights $\mathbf{w}$ of models in the model collection are loaded and reshaped into 2D matrices per layer, then divided into fixed-size tokens $\mathbf{T}_n$ of size $d_t$. Zero padding or splitting is applied where needed to ensure uniform dimensions, and a binary mask $\mathbf{M}$ is included to distinguish real parameters from padding. Each token is augmented with a 3D positional embedding $\mathbf{P}=[n,l,k]$ indicating absolute sequence position $n$, layer index $l$, and within-layer position $k$. For training, the tokenized model weights are divided into fixed-length chunks of a specified size (referred to as window) to allow uniform batch sizes and to efficiently process architectures of varying parameter counts. This representation allows processing models of different sizes and architectures in a unified sequence format. For simplicity, we drop the sequence indices $n$ in the following.

\vspace{-0.25cm}
\paragraph{Backbone and Learning Objective.}
The backbone itself operates as a sequence-to-sequence autoencoder with its bottleneck serving as the shared latent representation of model weights. It consists of an encoder $g_\theta$ that maps the input token sequence to a sequence of latent embeddings, $\mathbf{Z} = g_\theta(\mathbf{T}, \mathbf{P})$, and a decoder $h_\psi$ that reconstructs the original tokens from the latent embeddings, $ \widehat{\mathbf{T}} = h_\psi(\mathbf{Z}, \mathbf{P}).$
To structure the embedding space, a projection head $p_\phi$ maps the latent embeddings to a lower-dimensional space, $\mathbf{z}_p = p_\phi(\mathbf{Z})$,  which is used in a contrastive learning objective. Training is performed on chunks of token sequences with a combination of reconstruction and contrastive loss:
\begin{align}
    \mathcal{L}_{rec} &= \| \mathbf{M} \odot (\mathbf{T} - \widehat{\mathbf{T}} ) \|_2^2, \\
    \mathcal{L}_{c} &= NTXent(\mathbf{z}_{p,i}, \mathbf{z}_{p,j}), \\
    \mathcal{L} &= (1 - \gamma)\mathcal{L}_{rec} + \gamma \mathcal{L}_{c}.
\end{align}

\begin{table}[t]
\centering
\caption{Comparison between training from scratch and finetuning a GeoSANE-generated model across all benchmark datasets. We report accuracy for single-label classification, 
mAP for multi-label classification, 
mIoU for segmentation and mAP@0.5 for object detection. Best results are in \textbf{bold}. $\Delta$ indicates the absolute improvement over training from scratch.}
\small
\label{tab:scratch_vs_geosane}
\scalebox{0.9}{
\begin{tabular}{llccl}
\toprule
Dataset & Backbone & from scratch & \textbf{GeoSANE} & \multicolumn{1}{c}{\small $\Delta$} \\
\midrule
EuroSAT~\cite{Helber2017EuroSATAN}   & ViT-L & 95.0  & \textbf{99.1} & \textcolor{green!60!black}{+4.1} \\
RESISC-45~\cite{Cheng2017RemoteSI}   & ViT-L & 78.0  & \textbf{96.5} & \textcolor{green!60!black}{+18.5} \\
fMoW~\cite{Christie2017FunctionalMO}  & ViT-L & 35.7  & \textbf{58.9} & \textcolor{green!60!black}{+23.2} \\
Sen12Flood~\cite{rambour2020sen12}   & ViT-L & 80.4  & \textbf{85.2} & \textcolor{green!60!black}{+4.8} \\
Cal. Wildfires~\cite{california_fire_perimeters}  & ViT-L & 88.7  & \textbf{94.9} & \textcolor{green!60!black}{+6.2} \\
BigEarthNet~\cite{Sumbul2019BigearthnetAL}   & ViT-L & 69.8  & \textbf{88.7} & \textcolor{green!60!black}{+18.9} \\
\midrule
DFC2020~\cite{rha7-m332-19}    & Swin-B & 46.8  & \textbf{54.3} & \textcolor{green!60!black}{+7.5} \\
Spacenet1~\cite{Etten2018SpaceNetAR}      & Swin-B & 72.2  & \textbf{78.2} & \textcolor{green!60!black}{+6.0} \\
Sen1Floods11~\cite{Bonafilia2020Sen1Floods11AG}     & Swin-B & 81.0  & \textbf{89.6} & \textcolor{green!60!black}{+8.6} \\
\midrule
DIOR~\cite{Li2019ObjectDI}        & Swin-B & 67.5  & \textbf{79.0} & \textcolor{green!60!black}{+11.5} \\
\bottomrule
\end{tabular}}
\end{table}

\begin{table*}[t]
\centering
\caption{Comparison with existing remote sensing foundation models on multiple remote sensing benchmarks. 
We report Overall Accuracy for single-label classification, 
mean Average Precision for multi-label classification, mean intersection over union for segmentation and mean Average Precision (mAP@0.5) for object detection. For GeoSANE, we report the mean $\pm$ standard deviation over three independently generated models per prompt (Section \ref{sec:methods:geosane-generation}); for all baselines, we report the values reported in the respective papers.
Best results are in \textbf{bold}, second best are \underline{underlined}. $\Delta$ indicates the absolute improvement over the best baseline}
\label{tab:sota_comparison}
\setlength{\tabcolsep}{3pt} 
\scalebox{0.8}{
\begin{tabular}{llcccccccccc}
\toprule
\multirow{2}{*}{Model} & \multirow{2}{*}{Backbone} &
\multicolumn{5}{c}{Single-label} & 
\multicolumn{1}{c}{Multi-label} & \multicolumn{3}{c}{Segmentation } & \multicolumn{1}{c}{Object Det.} \\
\cmidrule(lr){3-7} \cmidrule(lr){8-8} \cmidrule(lr){9-11} \cmidrule(lr){12-12}
 & & EuroSAT  & RESISC45  & fMoW & Sen12Flood & Wildfires & BigEarthNet  & DFC2020 & SpaceNet & Sen1Floods11 & DIOR\\
\midrule
SatMAE \small  \cite{Cong2022SatMAEPT}  & ViT-L & 98.9 & 94.8 & 58.2  & 80.3 & 88.6 & 86.2 &44.1 & 78.1 & -- & 70.9\\
Scale-MAE \small \cite{Reed2022ScaleMAEAS} &  ViT-L & 99.1 & 95.7 & -- & 82.6 & 90.8 & 87.9  & -- & \textbf{78.9} & 74.1 & 73.8\\
RingMo \small \cite{Sun2023RingMoAR}    & Swin-B &  -- & 95.7 & --  & -- & -- &  --  & -- & -- & -- & 75.9\\
CROMA  \small \cite{Fuller2023CROMARS} &  ViT-L & \textbf{99.4} & -- & \textbf{59.0}  & \underline{83.4} & \underline{93.3} &  88.3 & \underline{49.8} & -- & \textbf{90.9} & --\\
GFM \small \cite{Mendieta2023TowardsGF} & Swin-B & -- & --  & -- & 77.9 & 91.9 & 86.3 & --& -- &72.6 & -- \\
SkySense \small \cite{Guo2023SkySenseAM} &  ViT-L & -- &  -- & -- &-- & --& \underline{88.6}  & -- &-- &-- & \underline{78.7} \\
MAPEX \small \cite{Mapex2026Hanna} & ViT-B & -- & -- & -- & 83.2 & 90.5 & -- & -- & -- & -- & -- \\
DOFA  \small \cite{Xiong2024NeuralPM} &  ViT-L & -- & \textbf{97.3} & -- & --& --& --  & --& -- & 89.4& --\\
\midrule
\textbf{GeoSANE} & \footnotesize ViT-L or Swin-B
  & \underline{99.1}\scriptsize$\pm$0.2 & \underline{96.5} \scriptsize$\pm$0.1 & \underline{58.9}\scriptsize$\pm$0.1 & \textbf{85.2}\scriptsize$\pm$0.3 & \textbf{94.9}\scriptsize$\pm$0.2 & \textbf{88.7}\scriptsize$\pm$0.1 & \textbf{54.3}\scriptsize$\pm$0.3& \underline{78.2}\scriptsize$\pm$0.3 & \underline{89.6}\scriptsize$\pm$0.1 & \textbf{79.0}\scriptsize$\pm$0.2 \\ 
\multicolumn{1}{c}{\small $\Delta$} &  & \small \textcolor{red!80!black}{-0.3} & \small \textcolor{red!80!black}{-0.8} &  \small \textcolor{red!80!black}{-0.1} & \small \textcolor{green!60!black}{+1.8} & \small \textcolor{green!60!black}{+1.6} & 
\small \textcolor{green!60!black}{+0.1} & 
\small \textcolor{green!60!black}{+4.5} & 
\small \textcolor{red!80!black}{-0.7} &
\small \textcolor{red!80!black}{-1.3} & 
\small \textcolor{green!60!black}{+0.3} \\
\bottomrule
\end{tabular}}
\end{table*}

Here, the mask $\mathbf{M}$ is used to separate real parameters with $1$ from padding with $0$, ensuring that the loss is only computed on actual weights. The contrastive term uses two augmented views $i,j$ of the same model: the first is the original token sequence, while the second is a noised version of it. Both views are processed through the encoder and projection head $p_{\phi}$, and the NT-Xent loss encourages their projected embeddings to be close in latent space while remaining distinct from embeddings of other models. To stabilize training,~\cite{Schrholt2024TowardsSA} required normalizing the weights of the models used for training layer-wise during pre-processing which limits the representation learning to models that share the same architecture. This limitation was addressed in recent work~\cite{Falk2025LearningMR} proposing to normalize the loss at runtime instead and therefore enabling learning weight-space representations of arbitrary models. Further, it demonstrated the feasibility of using models from publicly available model repositories such as Hugging Face as training data, instead of training the weight-space backbone on homogeneous populations of neural network models. 
This formulation is particularly important in our setting, where the goal is to learn from a broad collection of remote sensing models that vary across architectures, sensing modalities, and tasks. 



\subsection{Generating new Models from the Latent Space}
\label{sec:methods:geosane-generation}

Once the backbone is trained, GeoSANE can generate weights for new neural network models from the learned representation space. Given a \textit{prompt model} $a$ (e.g., an ImageNet-pretrained ViT-L or Swin-B from the \texttt{timm} library~\cite{rw2019timm}), we tokenize and encode its weights $\mathbf{w}_a$ into the latent space to obtain a latent representation $\mathbf{Z}_a = g_{\theta}(\mathbf{T}_a)$. We then fit a Kernel Density Estimator (KDE) around $\mathbf{Z}_a$ and draw samples $\tilde{\mathbf{z}}$ from this local distribution. This sampling procedure explores nearby regions in the latent space that are structurally similar to the prompt, while being shaped by the geospatial knowledge captured during training.
Each sampled latent representation $\tilde{\mathbf{z}}$ is decoded using the decoder to produce synthetic weight tokens $\tilde{\mathbf{T}} = h_{\psi}(\tilde{\mathbf{z}})$, which are subsequently de-tokenized into neural network weights $\tilde{\mathbf{w}}$. The resulting neural network shares the architecture of the \textit{prompt model} but differs in the actual parameter values, producing a network that is ready for fine-tuning on the target downstream task. Sampling in latent space is inexpensive, as both sampling and decoding require only forward passes. This makes it feasible to generate multiple candidate models and select top-$m$ candidates according to a simple performance criterion before fine-tuning.
\\
In practice, at inference, we use ViT-L prompts for classification and Swin-B prompts for segmentation and detection. We generate 10 candidate models per prompt and retain the best $m$=3 for fine-tuning. In Table~\ref{tab:geosane_multitask}, we evaluate the generation process using a larger set of prompt models, demonstrating that the method generalizes across backbones.

\section{Downstream Tasks and Datasets}
\label{sec:datasets}

\begin{table*}[t]
\centering
\small
\caption{Comparison of GeoSANE with pruning and distillation baselines. For \textit{Magnitude Pruning} and \textit{Variational Dropout}, we prune pretrained Remote Sensing Foundation Models (RSFMs) (ScaleMAE~\cite{Reed2022ScaleMAEAS} and SatMAE~\cite{Cong2022SatMAEPT}) and an ImageNet(IN)-pretrained ViT-L
 to obtain versions with approximately 11M non-zero parameters (ResNet-18) and 5M (MobileNetV2) non-zero parameters. For \textit{Knowledge Distillation}, the same RSFMs and the ImageNet ViT-L act as teachers, and the student networks are ResNet-18 models with 11M parameters or MobileNetV2 with 3.5M parameters. GeoSANE directly generates models of the target architecture and size. Best results are in \textbf{bold}, second best are \underline{underlined}. $\Delta$ indicates the absolute improvement over the best baseline}
 \scalebox{0.83}{
\begin{tabular}{ccccccccccccl}
\toprule
& \multirow{3}{*}{Dataset}
 & \multicolumn{3}{c}{Magnitude Pruning (MP)} & \multicolumn{3}{c}{Variational Dropout (VP)} &
\multicolumn{3}{c}{Knowledge Distillation (KD)} &
\multirow{3}{*}{\textbf{GeoSANE}} & \multirow{3}{*}{$\Delta$} \\
& & \multicolumn{3}{c}{Initial Models} & \multicolumn{3}{c}{Initial Models} & \multicolumn{3}{c}{Teachers} & & \\
\cmidrule(lr){3-5} \cmidrule(lr){6-8} \cmidrule(lr){9-11} 
&  & ViT-L \footnotesize (IN) & ScaleMAE & SatMAE &  ViT-L \footnotesize (IN) & ScaleMAE & SatMAE &
ViT-L \footnotesize (IN)& ScaleMAE & SatMAE & &
 \\
\midrule
\multirow{6}{*}{\rotatebox[origin=c]{90}{\small 11M Params}}  &
RESISC-45  & 88.1 & 87.2 & 84.9 & 81.1 & 81.8 & 80.6 & 90.2 & 90.3 & \underline{91.1} & \textbf{92.2} & \small \textcolor{green!60!black}{+1.1}\\
& EuroSAT   & 97.7  & 95.4 & 96.3 & 95.5 & 97.0 & 94.7 & \underline{97.9} & 94.0 & 94.7 & \textbf{98.7} &  \small \textcolor{green!60!black}{+0.8}\\
& fMoW & 33.7 & 36.2& 35.4& 25.2&32.3& 27.1& 38.6& \underline{43.1}& 41.9& \textbf{53.5} & \small \textcolor{green!60!black}{+10.4} \\
& BigEarthNet & 45.6 & 59.1 & 62.1 & 38.2 & 44.9 & 49.8 & 65.7 & \underline{67.3} & 66.5 & \textbf{83.7} & \textcolor{green!60!black}{+16.4} \\
& Sen12Flood & 77.2& 79.0& 76.5& 75.4& 75.3& 77.8& 79.3& \underline{82.1}& 80.8& \textbf{84.0} & \textcolor{green!60!black}{+1.9}\\
& Cal. Wildfires & 82.9& 82.7& 84.6& 79.1& 79.8& 81.3& 85.2& 87.6& \underline{88.0}& \textbf{91.6} & \textcolor{green!60!black}{+3.6}\\
\midrule
\multirow{6}{*}{\rotatebox[origin=c]{90}{\small 3.5M Params$^\ast$}}  &
RESISC-45  & 64.9& 65.6& 63.6& 61.2& 62.0& 61.0& 67.2 & 67.5 & \underline{68.1} & \textbf{70.0}  & \textcolor{green!60!black}{+1.9}\\
& EuroSAT   & 89.2 & 90.8& 91.3& 85.5 & 88.9& 90.7& 92.1 & 93.4 & \underline{94.8} & \textbf{96.2} & \textcolor{green!60!black}{+1.4} \\
& fMoW & 15.2 & 16.7& 20.5& 16.5& 17.8& 17.4& \underline{23.2}& \textbf{25.5}& 22.1& 17.7 & \textcolor{red!80!black}{-7.8}\\
& BigEarthNet & 42.5 & 41.8& 43.9& 32.8& 34.6& 37.1& 55.1  & \underline{60.7} & 58.5 & \textbf{73.3}  & \textcolor{green!60!black}{+12.6}\\
& Sen12Flood & 60.1& 62.7& 62.8& 55.9& 58.4& 59.5&62.1 & 62.8& \underline{64.7}& \textbf{70.2} & \textcolor{green!60!black}{+5.5}\\
& Cal. Wildfires & 68.0& 69.6& 69.9& 62.9& 63.7& 66.2& 72.9& 74.1& \underline{75.4}& \textbf{75.9} & \textcolor{green!60!black}{+0.5}\\
\bottomrule
\end{tabular}}
\raggedright
\footnotesize{$^\ast$ For MP and VP, models were pruned to approx. 5M parameters instead of 3.5M, as lower sparsity caused model instability.}
\label{tab:geosane_vs_prune_distill}
\vspace{-0.4cm}
\end{table*}

We evaluate models generated by GeoSANE on a diverse set of downstream tasks, covering classification, segmentation, and object detection across multiple modalities. 

\paragraph{Classification} We use six diverse datasets: RESISC45~\cite{Cheng2017RemoteSI} (RGB scene recognition), EuroSAT~\cite{Helber2017EuroSATAN} (multispectral land use classification), fMoW~\cite{Christie2017FunctionalMO} (multispectral scene classification, using 10\% of the training set following common practice~\cite{Maas2021SeasonalCU, Cong2022SatMAEPT}), BigEarthNet~\cite{Sumbul2019BigearthnetAL} (multispectral multi-label land cover classification), Sen12Flood~\cite{rambour2020sen12} (SAR-based flood detection), and California Wildfires~\cite{cal_fire_incidents, california_fire_perimeters} (SWIR-based wildfire detection).
We also evaluate on four classification benchmarks from GEO-Bench~\cite{Lacoste2023GEOBenchTF} (m-EuroSAT, m-BigEarthNet, m-So2Sat, m-Brick-Kiln) following~\cite{Danish2025TerraFMAS}, to assess the generalization on standardized remote sensing sets.
\vspace{-0.25cm}
\paragraph{Semantic Segmentation} We evaluate on three segmentation benchmarks: DFC2020~\cite{rha7-m332-19} (multispectral land cover mapping), Sen1Floods11~\cite{Bonafilia2020Sen1Floods11AG} (SAR-based flood segmentation) and SpaceNet1~\cite{Etten2018SpaceNetAR} (RGB building footprint extraction).
\vspace{-0.25cm}
\paragraph{Object Detection}
For object detection, we use the DIOR dataset~\cite{Li2019ObjectDI}, which contains 20 object categories in high-resolution RGB aerial imagery.
\vspace{0.15cm}

Together, these datasets cover a wide range of remote sensing tasks, multiple band types, and both single-label and multi-label setups, ensuring that the evaluation is comprehensive. Further dataset details are provided in the supplementary material.

\section{Experiments and Results}
\label{sec:experiments}

\subsection{Implementation Details}
\label{sec:experiments:pretraining_implementation}
GeoSANE is implemented as an autoencoder with approximately 900M parameters, where both the encoder and decoder are GPT-2 style transformers~\cite{Radford2019LanguageMA}. Model weights are reshaped and tokenized into fixed-size vectors of dimension 230, resulting in a total of approximately 165M tokens from our remote sensing model collection. To obtain a stronger latent representation, we first pretrain GeoSANE on a larger corpus of general computer vision models from HuggingFace~\cite{Falk2025LearningMR} (approximately 700M tokens), and then finetune on the remote sensing tokens. GeoSANE is trained for 150 epochs on a single NVIDIA H100 GPU. We use AdamW~\cite{Loshchilov2017DecoupledWD} with a learning rate of $2\times10^{-5}$, weight decay of $3\times10^{-9}$, and a OneCycleLR learning rate schedule. The model is optimized using a combination of reconstruction loss and contrastive guidance, as described in Section~\ref{sec:methods}. The checkpoint with the lowest validation loss is retained for downstream model generation.
For completeness, we also report in the supplementary material downstream results using models generated from the latent space \emph{before} fine-tuning GeoSANE on remote sensing data (i.e., pretrained only on general computer vision models).

\subsection{Performance of Generated Models}

We evaluate GeoSANE to test its ability to generate performant model weights across diverse remote sensing tasks and architectures.
Our experiments are designed to answer the following main questions:

\begin{enumerate}[label=\textbf{Q\arabic*.}, leftmargin=*, align=left]
    \item[\textbf{Q1.}] Does GeoSANE provide better initialization than training from scratch ?
    \item[\textbf{Q2.}] How do models generated by GeoSANE compare to existing remote sensing foundation models?
    \item[\textbf{Q3.}] Does GeoSANE go beyond model merging by learning relationships in weight space rather than just interpolating weights?
    \item[\textbf{Q4.}] Can GeoSANE generate strong lightweight models without explicit compression?
    \item[\textbf{Q5.}] How does GeoSANE improve and generalize across diverse model prompts?
\end{enumerate}

\paragraph{Experimental Setup } Unless otherwise specified, classification experiments use GeoSANE-generated ViT-L weights as the base architecture. For segmentation, we attach a lightweight segmentation head to a GeoSANE-generated Swin-B backbone. For object detection, following prior work~\cite{Sun2023RingMoAR, Guo2023SkySenseAM}, we use a Faster R-CNN detector~\cite{Ren2015FasterRT} with a Swin-B backbone as the feature extractor. \\
For downstream evaluation, all generated models are fine-tuned for 50 epochs\footnote{We finetune all models for 50 epochs for fair comparison, although performance usually saturates earlier (Fig~\ref{fig:conv_rates}).} using AdamW as the optimizer. We select the final checkpoint based on the lowest validation loss and report its corresponding test performance.

\subsubsection*{Q1: Initialization and Fine-tuning Performance}

Table \ref{tab:scratch_vs_geosane} compares models generated by GeoSANE with randomly initialized ones of identical architectures and finetuned for the same number of epochs, under similar conditions.  
Across ten diverse datasets, GeoSANE consistently outperforms training from scratch, with particularly large gains on more challenging, heterogeneous datasets such as fMoW and BigEarthNet, which contain many classes and fine-grained labels. These results show that GeoSANE can serve as an effective initializer for remote sensing models, across various modalities.

\subsubsection*{Q2: Comparison with Existing RSFMs}
Next, we benchmark GeoSANE against many exiting Remote Sensing Foundation Models including SatMAE~\cite{Cong2022SatMAEPT}, Scale-MAE~\cite{Reed2022ScaleMAEAS}, CROMA~\cite{Fuller2023CROMARS}, RingMo~\cite{Sun2023RingMoAR}, GFM~\cite{Mendieta2023TowardsGF}, SkySense ~\cite{Guo2023SkySenseAM}, MAPEX~\cite{Mapex2026Hanna} and DOFA~\cite{Xiong2024NeuralPM}. For classification tasks we use ViT-L backbones, and for segmentation and detection tasks we use Swin-B. As shown in Table \ref{tab:sota_comparison}, GeoSANE achieves the best or second-best results across ten benchmarks, matching or surpassing RSFMs.
These results show that GeoSANE can reach the same level of performance as models that rely on large-scale pretraining, while generating weights directly from its learned latent representation.

\subsubsection*{Evaluation on GEO-Bench}
We further evaluate GeoSANE on GEO-Bench~\cite{Lacoste2023GEOBenchTF}, which provides standardized benchmarks for remote sensing foundation models across multiple modalities and resolutions. As shown in Table~\ref{tab:geo_bench_terrafm}, GeoSANE achieves the best or second-best performance across all four classification datasets.

\subsubsection*{Q3: Comparison with Model Merging Methods}
Since GeoSANE learns a latent representation from many pretrained models, it is natural to ask whether simpler parameter-space combination techniques could achieve similar performances. A common baseline is model merging, where weights from different networks are combined directly in parameter space.
We therefore merge pairs of remote sensing foundation models using the DARE (\textbf{D}rop \textbf{A}nd \textbf{RE}scale)~\cite{Yu2023LanguageMA} method and compare the resulting merged models to GeoSANE-generated models of the same architecture. As shown in Table~\ref{tab:merged-model}, GeoSANE achieves consistently higher performance,  confirming that learning to generate weights in latent space goes beyond parameter averaging.\looseness-1

\begin{table}[t]
\centering
\caption{Comparison of a model generated by GeoSANE against a merged model obtained by combining RSFMs  using DARE \cite{Yu2023LanguageMA}, and against individual RSFMs.}
\label{tab:merged-model}
\scalebox{0.95}{
\begin{tabular}{lccc}
\toprule
\multirow{2}{*}{Model} &
\multicolumn{2}{c}{Single-label} & 
Multi-label \\
\cmidrule(lr){2-3} \cmidrule(lr){4-4}
 & EuroSAT  & RESISC45  & BigEarthNet  \\
\midrule
SatMAE \small \cite{Cong2022SatMAEPT}   & 98.9 & 94.8 & 86.2 \\
Scale-MAE \small \cite{Reed2022ScaleMAEAS}& \textbf{99.1} &  \underline{95.7}   & \underline{87.9} \\
\midrule
Merged Model & 96.4 & 86.1 & 69.0 \\
\midrule
\textbf{GeoSANE}  & \textbf{99.1} & \textbf{96.5}   & \textbf{88.7} \\
\bottomrule
\end{tabular}}
\vspace{-1em}
\end{table}

\begin{table}[t]
\centering
\caption{Comparison between finetuning GeoSANE-generated models vs. direct finetuning of the models used as prompts for GeoSANE (ImageNet-pretrained ViT-L for classification and Swin-B for segmentation and detection). Both versions receive the same finetuning budget of 50 epochs.}
\label{tab:anchor-model}
\scalebox{0.9}{
\begin{tabular}{lccl}
\toprule
Dataset &
Model Prompt & 
\textbf{GeoSANE} & \multicolumn{1}{c}{\small $\Delta$}\\
\midrule
EuroSAT    & 97.8 & \textbf{99.1}  & \small \textcolor{green!60!black}{+1.3}\\
RESISC45 &  92.3   & \textbf{96.5}  & \small \textcolor{green!60!black}{+4.2}\\
fMoW &  52.4   & \textbf{58.9}  & \small \textcolor{green!60!black}{+6.5}\\
Sen12Flood &  84.2   & \textbf{85.2}  & \small \textcolor{green!60!black}{+1.0} \\
Cal. Wildfires &  93.5   & \textbf{94.9}  &\small \textcolor{green!60!black}{+1.4} \\
BigEarthNet &  82.6   & \textbf{88.7}  &\small \textcolor{green!60!black}{+6.1} \\
\midrule
DFC2020 &  47.9 & \textbf{54.3} & \small \textcolor{green!60!black}{+6.4} \\
Spacenet1 & 75.8  & \textbf{78.2}  &\small \textcolor{green!60!black}{+2.4} \\
Sen1Floods11 & 85.2 & \textbf{89.6} &\small \textcolor{green!60!black}{+4.4}  \\
\midrule
DIOR &  73.6 & \textbf{79.0} & \small \textcolor{green!60!black}{+5.4} \\
\bottomrule
\end{tabular}}
\vspace{-1em}
\end{table}


\begin{table*}[t]
\centering
\small
\caption{Results on four classification tasks from GEO-Bench \cite{Lacoste2023GEOBenchTF}. All models are trained for 50 epochs. The reported numbers are overall accuracy (OA). The \textit{m-bigearthnet} dataset is evaluated using mAP. For GeoSANE, we report the mean $\pm$ standard deviation over three independently generated models per prompt (Section \ref{sec:methods:geosane-generation}); for all baselines, we report the values reported in the respective paper \cite{Danish2025TerraFMAS}. Best results are in \textbf{bold}, second best are \underline{underlined}. $\Delta$ indicates the absolute improvement over the best baseline}
\begin{tabular}{llcccc}
\toprule
Method & Backbone & m-eurosat & m-bigearthnet & m-so2sat & m-brick-kiln  \\
\midrule
SatMAE \cite{Cong2022SatMAEPT}& ViT-L & 96.6 & 68.3 & 57.2 & 98.4  \\
CROMA \cite{Fuller2023CROMARS} & ViT-L & 96.6 & 71.9 & 60.6 & 98.7  \\
DOFA  \cite{Xiong2024NeuralPM} & ViT-L & 96.9 & 68.0 & 58.7 & 98.6  \\
Prithvi-EO 2.0 \cite{Szwarcman2024PrithviEO20AV}  & ViT-L & 96.5 & 69.0 & 54.6 & 98.6  \\
AnySat  \cite{Astruc2024AnySatOE} & ViT-B  & 95.9 & 70.3 & 51.8 & 98.6  \\
Galileo \cite{Tseng2025GalileoLG}  & ViT-B  & 97.7 & 70.7 & 63.3 & \underline{98.7}  \\
TerraFM   \cite{Danish2025TerraFMAS}    & ViT-L & \textbf{98.6} & \underline{73.1} & \underline{64.9} & \textbf{99.0}  \\
\midrule
\textbf{GeoSANE} & ViT-L &  \underline{97.7}\scriptsize$\pm$0.1 & \textbf{74.2}\scriptsize$\pm$0.3 & \textbf{65.7}\scriptsize$\pm$0.2 &  98.6\scriptsize$\pm$0.2\\
\multicolumn{1}{c}{\small $\Delta$} &  & \small \textcolor{red!80!black}{-0.9} & \small \textcolor{green!60!black}{+1.1} & \small \textcolor{green!60!black}{+0.8} & \small \textcolor{red!80!black}{-0.4} \\
\bottomrule
\end{tabular}

\label{tab:geo_bench_terrafm}
\end{table*}

\subsubsection*{Q4: Comparison with Pruning and Distillation for Lightweight Model Generation}
A key advantage of GeoSANE is its ability to generate models of different sizes directly in weight space. This is particularly useful in remote sensing, where deployment often requires small and efficient models for on-board processing. To evaluate this, we compare GeoSANE to traditional compression techniques: magnitude pruning~\cite{Han2015LearningBW}, variational dropout~\cite{Molchanov2017VariationalDS}, and knowledge distillation~\cite{Hinton2015DistillingTK}. For pruning and variational dropout, we start from pretrained RSFMs and compress them to ResNet-18 (11M parameters) and MobileNetV2 (3.5M parameters) equivalents, followed by fine-tuning under the same training settings as GeoSANE. For distillation, the same RSFMs serve as teacher models, while ResNet-18 and MobileNetV2 act as students. In contrast, GeoSANE directly generates ResNet-18 and MobileNetV2 weights of the same parameter budgets, requiring neither a large teacher nor iterative pruning.
As shown in Table~\ref{tab:geosane_vs_prune_distill}, GeoSANE consistently outperforms both pruning and distillation baselines across all datasets and parameter budgets. These results demonstrate that GeoSANE can effectively generate lightweight, high-performing models directly, without relying on compression pipelines or teacher supervision.


\subsubsection*{Q5: Comparison With Prompt Models and Diversity of Generated Models}

\paragraph{GeoSANE-generated models vs.\ their prompts.} We next evaluate whether GeoSANE-generated models improve over the pretrained models used as prompts during generation.
For each task, we use an ImageNet-pretrained model as the prompt (ViT-L for classification, Swin-B for segmentation and detection) and generate new weights through GeoSANE.
We then fine-tune both the prompt and the generated model under identical settings.
As shown in Table~\ref{tab:anchor-model}, GeoSANE consistently yields higher performance, showing that the generation process produces meaningful and effective weight configurations. Since GeoSANE is trained on a diverse collection of remote sensing models, it learns weight patterns that are more aligned with geospatial data, enabling it to produce initializations that outperform ImageNet-pretrained prompts.

\begin{figure}
    \centering
    \includegraphics[width=\linewidth]{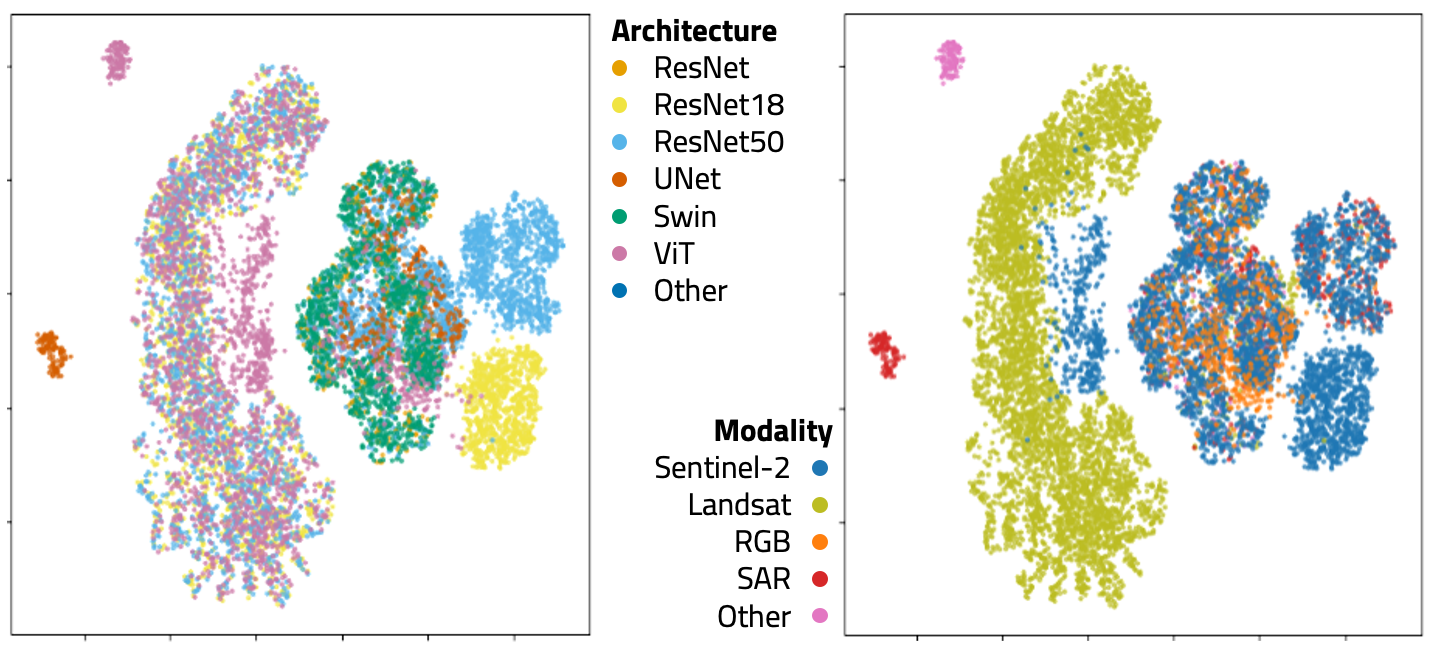}
    \caption{UMAPs Visualization of the latent weight space of GeoSANE, colored by architecture (left) and modality (right). GeoSANE learns a compact shared latent representation of model weights from neural network models of different architecture and modalities (see Sec. \ref{sec:supp:umap}, in the supplementary material for details).}
    \label{fig:umaps_horizontal}
    \vspace{-0.5cm}
\end{figure}

\begin{table}[t]
\centering
\small
\setlength{\tabcolsep}{6pt}
\renewcommand{\arraystretch}{1.15}

\caption{
We evaluate diverse GeoSANE-generated models on three tasks. All generated models are fine-tuned for 50 epochs. Please note, all models (for (a), (b), (c)) are generated using the same learned shared latent representation.
}

\textbf{(a) Classification with GeoSANE-generated Models} \\
\vspace{1mm}
\scalebox{0.92}{
\begin{tabular}{lcccc}
\toprule
\multirow{2}{*}{Dataset} & MobileNet & ResNet-18 & ViT-Base & ViT-Large \\
\cmidrule(lr){2-5}
& 3.5M & 11M & 86M & 300M \\
\midrule
RESISC-45   & 70.0  & 92.2 & 91.4 & 96.5 \\
EuroSAT    & 96.2  & 98.7 & 98.4 & 99.1 \\
fMoW        & 17.7  & 53.5 & 56.7 & 58.9 \\
BigEarthNet  &  73.3  & 83.7 & 86.9 & 88.7 \\
Sen12Flood  & 70.2 & 84.0 & 83.1 & 85.2 \\
Cal. Wildfires  & 75.9 & 91.6 & 93.8 & 94.9 \\
\bottomrule
\end{tabular}}

\vspace{3mm}

\textbf{(b) Semantic Segmentation with GeoSANE-generated Models} \\
\vspace{1mm}
\scalebox{0.92}{
\begin{tabular}{lcc}
\toprule
\multirow{2}{*}{Dataset} & UNet & Swin-Base \\
\cmidrule(lr){2-3}
& 17M & 88M \\
\midrule
DFC2020         & 48.3 & 54.3 \\
Spacenet1       & 76.7 & 78.2 \\
Sen1Floods11         & 84.2 & 89.2 \\
\bottomrule
\end{tabular}}

\vspace{3mm}

\textbf{(c) Object Detection with GeoSANE-generated Models} \\
\vspace{1mm}
\scalebox{0.92}{
\begin{tabular}{lccc}
\toprule
\multirow{2}{*}{Dataset} & ResNet-50 & Swin-Base & ViT-Large \\
\cmidrule(lr){2-4}
& 26M & 88M & 300M\\
\midrule
DIOR & 57.9 & 79.0 & 77.4
 \\
\bottomrule
\end{tabular}}

\label{tab:geosane_multitask}
\vspace{-0.5cm}
\end{table}


\paragraph{On-demand Diverse Models Generation}
Beyond matching the performance of existing RSFMs, GeoSANE also offers a major practical advantage: is not limited to a single backbone family.
It can generate weights for a diverse range of architectures and model sizes, including CNNs (ResNet, MobileNet, UNet, YOLO) and Transformers (ViT, Swin).
Across classification, segmentation, and detection tasks, these generated models show strong performance, proving that GeoSANE can generalize across very different architectures and tasks.
Table~\ref{tab:geosane_multitask} summarizes the results for all tasks and backbones.
All models are generated by GeoSANE and fine-tuned for 50 epochs.

\section{Conclusion}
In this work, we introduced GeoSANE, a model foundry that learns geospatial representations directly from the weights of existing models rather than from raw satellite data. By embedding a heterogeneous population of pretrained remote sensing (foundation) models into a shared latent representation, GeoSANE enables the generation of new model weights on demand, tailored to specific architectures and tasks. 

Across classification, segmentation, and detection tasks, GeoSANE consistently improves downstream performance over training from scratch, matches or surpasses state-of-the-art remote sensing foundation models, and is able to generate lightweight networks that outperform pruning- and distillation-based approaches. These results demonstrate that weight-space model generation is a competitive and effective alternative to large-scale pretraining. By learning from models instead of data, GeoSANE offers a scalable path to unify and transfer geospatial knowledge as the diversity and volume of remote sensing models continue to expand.\looseness-1
\label{sec:conclusion}

\section{Acknowledgments}
J.Hanna, D.Falk, and D.Borth would like to acknowledge the Swiss National Science Foundation (SNSF projects 213064 and 10001118), SPRIND (project Model Foundry), and the European Space Agency (ESA Phi-Lab CIN) for partial funding of this work. 
%
This work was also supported in part by the US National Science Foundation to S. Yu under awards NSF 2215542 and NSF 2313151.
\label{sec:acknowledgement}
{
    \small
    \bibliographystyle{ieeenat_fullname}

\begin{thebibliography}{63}
\providecommand{\natexlab}[1]{#1}
\providecommand{\url}[1]{\texttt{#1}}
\expandafter\ifx\csname urlstyle\endcsname\relax
  \providecommand{\doi}[1]{doi: #1}\else
  \providecommand{\doi}{doi: \begingroup \urlstyle{rm}\Url}\fi

\bibitem[Astruc et~al.(2024)Astruc, Gonthier, Mallet, and Landrieu]{Astruc2024AnySatOE}
Guillaume Astruc, Nicolas Gonthier, Cl{\'e}ment Mallet, and Loic Landrieu.
\newblock Anysat: One earth observation model for many resolutions, scales, and modalities.
\newblock \emph{2025 IEEE/CVF Conference on Computer Vision and Pattern Recognition (CVPR)}, pages 19530--19540, 2024.

\bibitem[Bonafilia et~al.(2020)Bonafilia, Tellman, Anderson, and Issenberg]{Bonafilia2020Sen1Floods11AG}
Derrick Bonafilia, Beth Tellman, Tyler Anderson, and Erica Issenberg.
\newblock Sen1floods11: a georeferenced dataset to train and test deep learning flood algorithms for sentinel-1.
\newblock \emph{2020 IEEE/CVF Conference on Computer Vision and Pattern Recognition Workshops (CVPRW)}, pages 835--845, 2020.

\bibitem[{California Department of Forestry and Fire Protection}({\natexlab{a}})]{cal_fire_incidents}
{California Department of Forestry and Fire Protection}.
\newblock {CAL FIRE Incidents}.
\newblock \url{https://www.fire.ca.gov/incidents}, {\natexlab{a}}.
\newblock Accessed: 2024-11.

\bibitem[{California Department of Forestry and Fire Protection}({\natexlab{b}})]{california_fire_perimeters}
{California Department of Forestry and Fire Protection}.
\newblock {California Fire Perimeters (all)}.
\newblock \url{https://catalog.data.gov/dataset/california-fire-perimeters-all-b3436}, {\natexlab{b}}.
\newblock Accessed: 2024-11.

\bibitem[Cheng et~al.(2017)Cheng, Han, and Lu]{Cheng2017RemoteSI}
Gong Cheng, Junwei Han, and Xiaoqiang Lu.
\newblock Remote sensing image scene classification: Benchmark and state of the art.
\newblock \emph{Proceedings of the IEEE}, 105:\penalty0 1865--1883, 2017.

\bibitem[Christie et~al.(2017)Christie, Fendley, Wilson, and Mukherjee]{Christie2017FunctionalMO}
Gordon~A. Christie, Neil Fendley, James Wilson, and Ryan Mukherjee.
\newblock Functional map of the world.
\newblock \emph{2018 IEEE/CVF Conference on Computer Vision and Pattern Recognition}, pages 6172--6180, 2017.

\bibitem[Cong et~al.(2022)Cong, Khanna, Meng, Liu, Rozi, He, Burke, Lobell, and Ermon]{Cong2022SatMAEPT}
Yezhen Cong, Samar Khanna, Chenlin Meng, Patrick Liu, Erik Rozi, Yutong He, Marshall Burke, David Lobell, and Stefano Ermon.
\newblock Satmae: Pre-training transformers for temporal and multi-spectral satellite imagery.
\newblock \emph{Advances in Neural Information Processing Systems}, 35:\penalty0 197--211, 2022.

\bibitem[Danish et~al.(2025)Danish, Munir, Shah, Khan, Anwer, Laaksonen, Khan, and Khan]{Danish2025TerraFMAS}
Muhammad~Sohail Danish, Muhammad~Akhtar Munir, Syed Roshaan~Ali Shah, Muhammad~Haris Khan, Rao~Muhammad Anwer, Jorma Laaksonen, Fahad~Shahbaz Khan, and Salman~H. Khan.
\newblock Terrafm: A scalable foundation model for unified multisensor earth observation.
\newblock \emph{ArXiv}, abs/2506.06281, 2025.

\bibitem[Drusch et~al.(2012)Drusch, Del~Bello, Carlier, Colin, Fernandez, Gascon, Hoersch, Isola, Laberinti, Martimort, et~al.]{drusch2012sentinel}
Matthias Drusch, Umberto Del~Bello, S{\'e}bastien Carlier, Olivier Colin, Veronica Fernandez, Ferran Gascon, Bianca Hoersch, Claudia Isola, Paolo Laberinti, Philippe Martimort, et~al.
\newblock Sentinel-2: Esa's optical high-resolution mission for gmes operational services.
\newblock \emph{Remote sensing of Environment}, 120:\penalty0 25--36, 2012.

\bibitem[Etten et~al.(2018)Etten, Lindenbaum, and Bacastow]{Etten2018SpaceNetAR}
Adam~Van Etten, David Lindenbaum, and Todd~M. Bacastow.
\newblock Spacenet: A remote sensing dataset and challenge series.
\newblock \emph{ArXiv}, abs/1807.01232, 2018.

\bibitem[Falk et~al.(2025{\natexlab{a}})Falk, Meynent, Pfammatter, Sch{\"u}rholt, and Borth]{falk2025vitzoo}
Damian Falk, L{\'e}o Meynent, Florence Pfammatter, Konstantin Sch{\"u}rholt, and Damian Borth.
\newblock A model zoo of vision transformers.
\newblock In \emph{Workshop on Neural Network Weights as a New Data Modality}, 2025{\natexlab{a}}.

\bibitem[Falk et~al.(2025{\natexlab{b}})Falk, Sch{\"u}rholt, Tzevelekakis, Meynent, and Borth]{Falk2025LearningMR}
Damian Falk, Konstantin Sch{\"u}rholt, Konstantinos Tzevelekakis, L{\'e}o Meynent, and Damian Borth.
\newblock Learning model representations using publicly available model hubs.
\newblock \emph{ArXiv}, abs/2510.02096, 2025{\natexlab{b}}.

\bibitem[Fuller et~al.(2023)Fuller, Millard, and Green]{Fuller2023CROMARS}
Anthony Fuller, Koreen Millard, and James Green.
\newblock Croma: Remote sensing representations with contrastive radar-optical masked autoencoders.
\newblock \emph{Advances in Neural Information Processing Systems}, 36:\penalty0 5506--5538, 2023.

\bibitem[Garioud et~al.(2023)Garioud, Gonthier, Landrieu, De~Wit, Valette, Poup{\'e}e, Giordano, et~al.]{Garioud2023FLAIRAC}
Anatol Garioud, Nicolas Gonthier, Loic Landrieu, Apolline De~Wit, Marion Valette, Marc Poup{\'e}e, S{\'e}bastien Giordano, et~al.
\newblock Flair: a country-scale land cover semantic segmentation dataset from multi-source optical imagery.
\newblock \emph{Advances in Neural Information Processing Systems}, 36:\penalty0 16456--16482, 2023.

\bibitem[Guo et~al.(2023)Guo, Lao, Dang, Zhang, Yu, Ru, Zhong, Huang, Wu, Hu, He, Wang, Chen, Yang, Zhang, and Li]{Guo2023SkySenseAM}
Xin Guo, Jiangwei Lao, Bo Dang, Yingying Zhang, Lei Yu, Lixiang Ru, Liheng Zhong, Ziyuan Huang, Kang Wu, Dingxiang Hu, Huimei He, Jian Wang, Jingdong Chen, Ming Yang, Yongjun Zhang, and Yansheng Li.
\newblock Skysense: A multi-modal remote sensing foundation model towards universal interpretation for earth observation imagery.
\newblock \emph{2024 IEEE/CVF Conference on Computer Vision and Pattern Recognition (CVPR)}, pages 27662--27673, 2023.

\bibitem[Ha et~al.(2017)Ha, Dai, and Le]{Ha2016HyperNetworks}
David Ha, Andrew~M Dai, and Quoc~V Le.
\newblock Hypernetworks.
\newblock In \emph{International Conference on Learning Representations}, 2017.

\bibitem[Han et~al.(2015)Han, Pool, Tran, and Dally]{Han2015LearningBW}
Song Han, Jeff Pool, John Tran, and William~J. Dally.
\newblock Learning both weights and connections for efficient neural network.
\newblock In \emph{Neural Information Processing Systems}, 2015.

\bibitem[Han et~al.(2026)Han, Wang, Zhao, Zhang, Li, Borth, Yu, Maron, Ye, Yin, et~al.]{han2026survey}
Xiaolong Han, Zehong Wang, Bo Zhao, Binchi Zhang, Jundong Li, Damian Borth, Rose Yu, Haggai Maron, Yanfang Ye, Lu Yin, et~al.
\newblock A survey of weight space learning: Understanding, representation, and generation.
\newblock \emph{arXiv preprint arXiv:2603.10090}, 2026.

\bibitem[Hanna and Borth(2025)]{hanna2025know}
Jo{\"e}lle Hanna and Damian Borth.
\newblock Know your attention maps: Class-specific token masking for weakly supervised semantic segmentation.
\newblock In \emph{Proceedings of the IEEE/CVF International Conference on Computer Vision}, pages 23763--23772, 2025.

\bibitem[Hanna et~al.(2026)Hanna, Scheibenreif, and Borth]{Mapex2026Hanna}
Joëlle Hanna, Linus Scheibenreif, and Damian Borth.
\newblock Mapex: Modality-aware pruning of experts for remote sensing foundation models.
\newblock \emph{IEEE Transactions on Geoscience and Remote Sensing}, 64:\penalty0 1--11, 2026.

\bibitem[He et~al.(2025)He, Zhou, Liu, Cao, and Ma]{he2025research}
Lu-hao He, Yong-zhang Zhou, Lei Liu, Wei Cao, and Jian-hua Ma.
\newblock Research on object detection and recognition in remote sensing images based on yolov11.
\newblock \emph{Scientific Reports}, 15\penalty0 (1):\penalty0 14032, 2025.

\bibitem[Helber et~al.(2017)Helber, Bischke, Dengel, and Borth]{Helber2017EuroSATAN}
Patrick Helber, Benjamin Bischke, Andreas~R. Dengel, and Damian Borth.
\newblock Eurosat: A novel dataset and deep learning benchmark for land use and land cover classification.
\newblock \emph{IEEE Journal of Selected Topics in Applied Earth Observations and Remote Sensing}, 12:\penalty0 2217--2226, 2017.

\bibitem[Hinton et~al.(2015)Hinton, Vinyals, and Dean]{Hinton2015DistillingTK}
Geoffrey~E. Hinton, Oriol Vinyals, and Jeffrey Dean.
\newblock Distilling the knowledge in a neural network.
\newblock \emph{ArXiv}, abs/1503.02531, 2015.

\bibitem[Horwitz et~al.(2025)Horwitz, Kurer, Kahana, Amar, and Hoshen]{horwitz2025we}
Eliahu Horwitz, Nitzan Kurer, Jonathan Kahana, Liel Amar, and Yedid Hoshen.
\newblock We should chart an atlas of all the world's models.
\newblock \emph{Advances in Neural Information Processing Systems}, 2025.

\bibitem[Jakubik et~al.(2023)Jakubik, Roy, Phillips, Fraccaro, Godwin, Zadrozny, Szwarcman, Gomes, Nyirjesy, Edwards, Kimura, Simumba, Chu, Mukkavilli, Lambhate, Das, Bangalore, Oliveira, Muszynski, Ankur, Ramasubramanian, Gurung, Khallaghi, Li, Cecil, Ahmadi, Kordi, Alemohammad, Maskey, Ganti, Weldemariam, and Ramachandran]{Prithvi-100M-preprint}
Johannes Jakubik, Sujit Roy, C.~E. Phillips, Paolo Fraccaro, Denys Godwin, Bianca Zadrozny, Daniela Szwarcman, Carlos Gomes, Gabby Nyirjesy, Blair Edwards, Daiki Kimura, Naomi Simumba, Linsong Chu, S.~Karthik Mukkavilli, Devyani Lambhate, Kamal Das, Ranjini Bangalore, Dario Oliveira, Michal Muszynski, Kumar Ankur, Muthukumaran Ramasubramanian, Iksha Gurung, Sam Khallaghi, Hanxi~(Steve) Li, Michael Cecil, Maryam Ahmadi, Fatemeh Kordi, Hamed Alemohammad, Manil Maskey, Raghu Ganti, Kommy Weldemariam, and Rahul Ramachandran.
\newblock {Foundation Models for Generalist Geospatial Artificial Intelligence}.
\newblock \emph{Preprint Available on arxiv:2310.18660}, 2023.

\bibitem[Jakubik et~al.(2025)Jakubik, Yang, Blumenstiel, Scheurer, Sedona, Maurogiovanni, Bosmans, Dionelis, Marsocci, Kopp, et~al.]{Jakubik2025TerraMindLG}
Johannes Jakubik, Felix Yang, Benedikt Blumenstiel, Erik Scheurer, Rocco Sedona, Stefano Maurogiovanni, Jente Bosmans, Nikolaos Dionelis, Valerio Marsocci, Niklas Kopp, et~al.
\newblock Terramind: Large-scale generative multimodality for earth observation.
\newblock In \emph{Proceedings of the IEEE/CVF International Conference on Computer Vision}, pages 7383--7394, 2025.

\bibitem[Lacoste et~al.(2023)Lacoste, Lehmann, Rodriguez, Sherwin, Kerner, L{\"u}tjens, Irvin, Dao, Alemohammad, Drouin, et~al.]{Lacoste2023GEOBenchTF}
Alexandre Lacoste, Nils Lehmann, Pau Rodriguez, Evan Sherwin, Hannah Kerner, Bj{\"o}rn L{\"u}tjens, Jeremy Irvin, David Dao, Hamed Alemohammad, Alexandre Drouin, et~al.
\newblock Geo-bench: Toward foundation models for earth monitoring.
\newblock \emph{Advances in Neural Information Processing Systems}, 36:\penalty0 51080--51093, 2023.

\bibitem[Li et~al.(2019)Li, Wan, Cheng, Meng, and Han]{Li2019ObjectDI}
Ke Li, Gang Wan, Gong Cheng, Liqiu Meng, and Junwei Han.
\newblock Object detection in optical remote sensing images: A survey and a new benchmark.
\newblock \emph{ArXiv}, abs/1909.00133, 2019.

\bibitem[Loshchilov and Hutter(2017)]{Loshchilov2017DecoupledWD}
Ilya Loshchilov and Frank Hutter.
\newblock Decoupled weight decay regularization.
\newblock In \emph{International Conference on Learning Representations}, 2017.

\bibitem[Lu et~al.(2024)Lu, Guo, Zimmer-Dauphinee, Nieusma, Wang, VanValkenburgh, Wernke, and Huo]{Lu2024VisionFM}
Siqi Lu, Junlin Guo, James Zimmer-Dauphinee, Jordan~M. Nieusma, Xiao Wang, Parker VanValkenburgh, Steven~A. Wernke, and Yuankai Huo.
\newblock Vision foundation models in remote sensing: A survey.
\newblock 2024.

\bibitem[Ma{\~n}as et~al.(2021)Ma{\~n}as, Lacoste, i~Nieto, V{\'a}zquez, and L{\'o}pez]{Maas2021SeasonalCU}
Oscar Ma{\~n}as, Alexandre Lacoste, Xavier~Gir{\'o} i Nieto, David V{\'a}zquez, and Pau~Rodr{\'i}guez L{\'o}pez.
\newblock Seasonal contrast: Unsupervised pre-training from uncurated remote sensing data.
\newblock \emph{2021 IEEE/CVF International Conference on Computer Vision (ICCV)}, pages 9394--9403, 2021.

\bibitem[Mendieta et~al.(2023)Mendieta, Han, Shi, Zhu, Chen, and Li]{Mendieta2023TowardsGF}
Matias Mendieta, Boran Han, Xingjian Shi, Yi Zhu, Chen Chen, and Mu Li.
\newblock Towards geospatial foundation models via continual pretraining.
\newblock \emph{2023 IEEE/CVF International Conference on Computer Vision (ICCV)}, pages 16760--16770, 2023.

\bibitem[Molchanov et~al.(2017)Molchanov, Ashukha, and Vetrov]{Molchanov2017VariationalDS}
Dmitry Molchanov, Arsenii Ashukha, and Dmitry~P. Vetrov.
\newblock Variational dropout sparsifies deep neural networks.
\newblock In \emph{International Conference on Machine Learning}, 2017.

\bibitem[Peebles et~al.(2022)Peebles, Radosavovic, Brooks, Efros, and Malik]{Peebles2022LearningTL}
William~S. Peebles, Ilija Radosavovic, Tim Brooks, Alexei~A. Efros, and Jitendra Malik.
\newblock Learning to learn with generative models of neural network checkpoints.
\newblock \emph{ArXiv}, abs/2209.12892, 2022.

\bibitem[Radford et~al.(2019)Radford, Wu, Child, Luan, Amodei, and Sutskever]{Radford2019LanguageMA}
Alec Radford, Jeff Wu, Rewon Child, David Luan, Dario Amodei, and Ilya Sutskever.
\newblock Language models are unsupervised multitask learners.
\newblock 2019.

\bibitem[Rambour et~al.(2020)Rambour, Audebert, Koeniguer, Le~Saux, Crucianu, and Datcu]{rambour2020sen12}
C Rambour, N Audebert, E Koeniguer, B Le~Saux, M Crucianu, and M Datcu.
\newblock {Sen12-flood: a SAR and Multispectral Dataset for Flood Detection}.
\newblock \emph{IEEE: Piscataway, NJ, USA}, 2020.

\bibitem[Reed et~al.(2022)Reed, Gupta, Li, Brockman, Funk, Clipp, Candido, Uyttendaele, and Darrell]{Reed2022ScaleMAEAS}
Colorado Reed, Ritwik Gupta, Shufan Li, Sara Brockman, Christopher Funk, Brian Clipp, Salvatore Candido, Matthew Uyttendaele, and Trevor Darrell.
\newblock Scale-mae: A scale-aware masked autoencoder for multiscale geospatial representation learning.
\newblock \emph{2023 IEEE/CVF International Conference on Computer Vision (ICCV)}, pages 4065--4076, 2022.

\bibitem[Ren et~al.(2015)Ren, He, Girshick, and Sun]{Ren2015FasterRT}
Shaoqing Ren, Kaiming He, Ross~B. Girshick, and Jian Sun.
\newblock Faster r-cnn: Towards real-time object detection with region proposal networks.
\newblock \emph{IEEE Transactions on Pattern Analysis and Machine Intelligence}, 39:\penalty0 1137--1149, 2015.

\bibitem[Scheibenreif et~al.(2022)Scheibenreif, Hanna, Mommert, and Borth]{scheibenreif2022self}
Linus Scheibenreif, Jo{\"e}lle Hanna, Michael Mommert, and Damian Borth.
\newblock {Self-supervised Vision Transformers for Land-cover Segmentation and Classification}.
\newblock In \emph{Proceedings of the IEEE/CVF Conference on Computer Vision and Pattern Recognition}, pages 1422--1431, 2022.

\bibitem[Schmitt et~al.(2019)Schmitt, Hughes, Ghamisi, Yokoya, and Hänsch]{rha7-m332-19}
Michael Schmitt, Lloyd Hughes, Pedram Ghamisi, Naoto Yokoya, and Ronny Hänsch.
\newblock 2020 ieee grss data fusion contest, 2019.

\bibitem[Sch{\"u}rholt et~al.(2021)Sch{\"u}rholt, Kostadinov, and Borth]{Schrholt2021HyperRepresentationsSR}
Konstantin Sch{\"u}rholt, Dimche Kostadinov, and Damian Borth.
\newblock Hyper-representations: Self-supervised representation learning on neural network weights for model characteristic prediction.
\newblock 2021.

\bibitem[Sch{\"u}rholt et~al.(2022{\natexlab{a}})Sch{\"u}rholt, Knyazev, Gir{\'o}-i Nieto, and Borth]{Schrholt2022HyperRepresentationsAG}
Konstantin Sch{\"u}rholt, Boris Knyazev, Xavier Gir{\'o}-i Nieto, and Damian Borth.
\newblock Hyper-representations as generative models: Sampling unseen neural network weights.
\newblock \emph{Advances in Neural Information Processing Systems}, 35:\penalty0 27906--27920, 2022{\natexlab{a}}.

\bibitem[Sch{\"u}rholt et~al.(2022{\natexlab{b}})Sch{\"u}rholt, Taskiran, Knyazev, Gir{\'o}-i Nieto, and Borth]{schurholt2022model}
Konstantin Sch{\"u}rholt, Diyar Taskiran, Boris Knyazev, Xavier Gir{\'o}-i Nieto, and Damian Borth.
\newblock Model zoos: A dataset of diverse populations of neural network models.
\newblock \emph{Advances in Neural Information Processing Systems}, 35:\penalty0 38134--38148, 2022{\natexlab{b}}.

\bibitem[Sch{\"u}rholt et~al.(2024{\natexlab{a}})Sch{\"u}rholt, Bouritsas, Horwitz, Lim, Gelberg, Zhao, Zhou, Borth, and Jegelka]{schurholt2024neural}
Konstantin Sch{\"u}rholt, Giorgos Bouritsas, Eliahu Horwitz, Derek Lim, Yoav Gelberg, Bo Zhao, Allan Zhou, Damian Borth, and Stefanie Jegelka.
\newblock Neural network weights as a new data modality.
\newblock In \emph{ICLR 2025 Workshop Proposals}, 2024{\natexlab{a}}.

\bibitem[Sch{\"u}rholt et~al.(2024{\natexlab{b}})Sch{\"u}rholt, Mahoney, and Borth]{Schrholt2024TowardsSA}
Konstantin Sch{\"u}rholt, Michael~W Mahoney, and Damian Borth.
\newblock Towards scalable and versatile weight space learning.
\newblock In \emph{International Conference on Machine Learning}, pages 43947--43966. PMLR, 2024{\natexlab{b}}.

\bibitem[Sch{\"u}rholt et~al.(2025)Sch{\"u}rholt, Meynent, Zhou, Lu, Yang, and Borth]{schurholt2025phasetransitions}
Konstantin Sch{\"u}rholt, L{\'e}o Meynent, Yefan Zhou, Haiquan Lu, Yaoqing Yang, and Damian Borth.
\newblock A model zoo on phase transitions in neural networks.
\newblock \emph{Journal of Data-centric Machine Learning Research}, 2025.

\bibitem[Soro et~al.()Soro, Andreis, Lee, Jeong, Chong, Hutter, and Hwang]{Soro2024DiffusionbasedNN}
Bedionita Soro, Bruno Andreis, Hayeon Lee, Wonyong Jeong, Song Chong, Frank Hutter, and Sung~Ju Hwang.
\newblock Diffusion-based neural network weights generation.
\newblock In \emph{The Thirteenth International Conference on Learning Representations}.

\bibitem[Stewart et~al.(2021)Stewart, Robinson, Corley, Ortiz, Ferres, and Banerjee]{Stewart2021TorchGeoDL}
Adam~J. Stewart, Caleb Robinson, Isaac~A. Corley, Anthony Ortiz, Juan M.~Lavista Ferres, and Arindam Banerjee.
\newblock Torchgeo: deep learning with geospatial data.
\newblock \emph{Proceedings of the 30th International Conference on Advances in Geographic Information Systems}, 2021.

\bibitem[Sumbul et~al.(2019)Sumbul, Charfuelan, Demir, and Markl]{Sumbul2019BigearthnetAL}
Gencer Sumbul, Marcela Charfuelan, Beg{\"u}m Demir, and Volker Markl.
\newblock Bigearthnet: A large-scale benchmark archive for remote sensing image understanding.
\newblock \emph{IGARSS 2019 - 2019 IEEE International Geoscience and Remote Sensing Symposium}, pages 5901--5904, 2019.

\bibitem[Sun et~al.(2023)Sun, Wang, Lu, Zhu, Lu, He, Li, Rong, Yang, Chang, He, Yang, Wang, Lu, and Fu]{Sun2023RingMoAR}
Xian Sun, Peijin Wang, Wanxuan Lu, Zicong Zhu, Xiaonan Lu, Qi He, Junxi Li, Xuee Rong, Zhujun Yang, Hao Chang, Qinglin He, Guang Yang, Ruiping Wang, Jiwen Lu, and Kun Fu.
\newblock Ringmo: A remote sensing foundation model with masked image modeling.
\newblock \emph{IEEE Transactions on Geoscience and Remote Sensing}, 61:\penalty0 1--22, 2023.

\bibitem[Szwarcman et~al.(2025)Szwarcman, Roy, Fraccaro, G{\'\i}slason, Blumenstiel, Ghosal, De~Oliveira, de~Sousa~Almeida, Sedona, Kang, et~al.]{Szwarcman2024PrithviEO20AV}
Daniela Szwarcman, Sujit Roy, Paolo Fraccaro, Orsteinn~El{\'\i} G{\'\i}slason, Benedikt Blumenstiel, Rinki Ghosal, Pedro~Henrique De~Oliveira, Joao~Lucas de Sousa~Almeida, Rocco Sedona, Yanghui Kang, et~al.
\newblock Prithvi-eo-2.0: A versatile multi-temporal foundation model for earth observation applications.
\newblock \emph{IEEE Transactions on Geoscience and Remote Sensing}, 2025.

\bibitem[Torres et~al.(2012)Torres, Snoeij, Geudtner, Bibby, Davidson, Attema, Potin, Rommen, Floury, Brown, et~al.]{torres2012gmes}
Ramon Torres, Paul Snoeij, Dirk Geudtner, David Bibby, Malcolm Davidson, Evert Attema, Pierre Potin, Bj{\"O}rn Rommen, Nicolas Floury, Mike Brown, et~al.
\newblock Gmes sentinel-1 mission.
\newblock \emph{Remote sensing of environment}, 120:\penalty0 9--24, 2012.

\bibitem[Tseng et~al.(2025)Tseng, Fuller, Reil, Herzog, Beukema, Bastani, Green, Shelhamer, Kerner, and Rolnick]{Tseng2025GalileoLG}
Gabriel Tseng, Anthony Fuller, Marlena Reil, Henry Herzog, Patrick Beukema, Favyen Bastani, James~R. Green, Evan Shelhamer, Hannah Kerner, and David Rolnick.
\newblock Galileo: Learning global\&local features of many remote sensing modalities.
\newblock 2025.

\bibitem[Wang et~al.(2024)Wang, Xu, Zhou, Zang, Darrell, Liu, and You]{Wang2024NeuralND}
Kaitian Wang, Zhaopan Xu, Yukun Zhou, Zelin Zang, Trevor Darrell, Zhuang Liu, and Yang You.
\newblock Neural network diffusion.
\newblock \emph{ArXiv}, abs/2402.13144, 2024.

\bibitem[Wang et~al.(2025)Wang, Tang, Zhao, Sch{\"u}rholt, Wang, and You]{wang2025scaling}
Kai Wang, Dongwen Tang, Wangbo Zhao, Konstantin Sch{\"u}rholt, Zhangyang Wang, and Yang You.
\newblock Scaling up parameter generation: A recurrent diffusion approach.
\newblock In \emph{The Thirty-ninth Annual Conference on Neural Information Processing Systems}, 2025.

\bibitem[Wang et~al.(2023)Wang, Braham, Xiong, Liu, Albrecht, and Zhu]{Wang2022SSL4EOS12AL}
Yi Wang, Nassim Ait~Ali Braham, Zhitong Xiong, Chenying Liu, Conrad~M Albrecht, and Xiao~Xiang Zhu.
\newblock Ssl4eo-s12: A large-scale multimodal, multitemporal dataset for self-supervised learning in earth observation [software and data sets].
\newblock \emph{IEEE Geoscience and Remote Sensing Magazine}, 11\penalty0 (3):\penalty0 98--106, 2023.

\bibitem[Wightman(2019)]{rw2019timm}
Ross Wightman.
\newblock Pytorch image models.
\newblock \url{https://github.com/rwightman/pytorch-image-models}, 2019.

\bibitem[Wortsman et~al.(2021)Wortsman, Ilharco, Li, Kim, Hajishirzi, Farhadi, Namkoong, and Schmidt]{Wortsman2021RobustFO}
Mitchell Wortsman, Gabriel Ilharco, Mike Li, Jong~Wook Kim, Hannaneh Hajishirzi, Ali Farhadi, Hongseok Namkoong, and Ludwig Schmidt.
\newblock Robust fine-tuning of zero-shot models.
\newblock \emph{2022 IEEE/CVF Conference on Computer Vision and Pattern Recognition (CVPR)}, pages 7949--7961, 2021.

\bibitem[Wortsman et~al.(2022)Wortsman, Ilharco, Gadre, Roelofs, Gontijo-Lopes, Morcos, Namkoong, Farhadi, Carmon, Kornblith, et~al.]{Wortsman2022ModelSA}
Mitchell Wortsman, Gabriel Ilharco, Samir~Ya Gadre, Rebecca Roelofs, Raphael Gontijo-Lopes, Ari~S Morcos, Hongseok Namkoong, Ali Farhadi, Yair Carmon, Simon Kornblith, et~al.
\newblock Model soups: averaging weights of multiple fine-tuned models improves accuracy without increasing inference time.
\newblock In \emph{International conference on machine learning}, pages 23965--23998. PMLR, 2022.

\bibitem[Xiao et~al.(2025)Xiao, Xuan, Wang, Huang, Tao, Lu, and Yokoya]{Xiao2024FoundationMF}
Aoran Xiao, Weihao Xuan, Junjue Wang, Jiaxing Huang, Dacheng Tao, Shijian Lu, and Naoto Yokoya.
\newblock Foundation models for remote sensing and earth observation: A survey.
\newblock \emph{IEEE Geoscience and Remote Sensing Magazine}, 2025.

\bibitem[Xiong et~al.(2024)Xiong, Wang, Zhang, Stewart, Hanna, Borth, Papoutsis, Saux, Camps-Valls, and Zhu]{Xiong2024NeuralPM}
Zhitong Xiong, Yi Wang, Fahong Zhang, Adam~J. Stewart, Joelle Hanna, Damian Borth, Ioannis Papoutsis, Bertrand~Le Saux, Gustau Camps-Valls, and Xiao~Xiang Zhu.
\newblock Neural plasticity-inspired multimodal foundation model for earth observation.
\newblock 2024.

\bibitem[Yadav et~al.(2023)Yadav, Tam, Choshen, Raffel, and Bansal]{Yadav2023TIESMergingRI}
Prateek Yadav, Derek Tam, Leshem Choshen, Colin Raffel, and Mohit Bansal.
\newblock Ties-merging: Resolving interference when merging models.
\newblock In \emph{Neural Information Processing Systems}, 2023.

\bibitem[Yu et~al.(2024)Yu, Yu, Yu, Huang, and Li]{Yu2023LanguageMA}
Le Yu, Bowen Yu, Haiyang Yu, Fei Huang, and Yongbin Li.
\newblock Language models are super mario: Absorbing abilities from homologous models as a free lunch.
\newblock In \emph{International Conference on Machine Learning}, pages 57755--57775. PMLR, 2024.

\end{thebibliography}

}

\clearpage
\setcounter{page}{1}
\maketitlesupplementary


\section{Effect of Remote Sensing Fine-tuning on GeoSANE Model Generation.}
To assess the contribution of remote sensing–specific fine-tuning in GeoSANE’s latent weight space, we compare models generated from two variants of the backbone: (1) \texttt{CV-only}, where GeoSANE is trained only on general computer vision models~\cite{Falk2025LearningMR}, and (2) \texttt{CV+RS}, where GeoSANE is additionally finetuned on our remote sensing (RS) model collection. For each setting, we generate weights for the same target architectures (ViT-L) and finetune them identically on downstream tasks. This allows us to isolate the effect of training GeoSANE on remote sensing weights on the quality of the generated models. Results are shown in Table~\ref{tab:cv_vs_rs_finetune}.\\ \indent
When comparing Table~\ref{tab:cv_vs_rs_finetune} with the results in Table~\ref{tab:anchor-model} (models prompt vs. GeoSANE-generated models), we observe that models generated without remote sensing fine-tuning (\texttt{CV-only}) perform similarly to their ImageNet-pretrained model prompts. Since these model prompts are themselves ImageNet models, the latent space trained only on computer vision models provides limited additional benefit. In contrast, with additional training on remote sensing model weights \texttt{CV+RS}, GeoSANE-generated models clearly outperform their model prompts. This demonstrates that training on remote sensing model weights is essential: it injects geospatial structure into the latent space and enables the generation of initializations that outperform ImageNet-pretrained prompts.

\begin{table}[h]
\centering
\caption{Downstream performance of models generated by GeoSANE trained without (\texttt{CV-only}) and with (\texttt{CV+RS}) remote sensing model weights.}
\label{tab:cv_vs_rs_finetune}
\scalebox{0.9}{
\begin{tabular}{lccl}
\toprule
Dataset &
\texttt{~~~CV-only~~~} & 
\texttt{~~~CV+RS~~~} & \multicolumn{1}{c}{\small $\Delta$}\\
\midrule
EuroSAT    &  97.8 & \textbf{99.1}  & \small \textcolor{green!60!black}{+1.3}\\
RESISC45 &   93.4  & \textbf{96.5}  & \small \textcolor{green!60!black}{+3.1}\\
fMoW &   53.1  & \textbf{58.9}  & \small \textcolor{green!60!black}{+5.8}\\
Sen12Flood &   82.9  & \textbf{85.2}  & \small \textcolor{green!60!black}{+2.3} \\
Cal. Wildfires &   92.2  & \textbf{94.9}  &\small \textcolor{green!60!black}{+2.7} \\
BigEarthNet &  83.0   & \textbf{88.7}  &\small \textcolor{green!60!black}{+5.7} \\
\midrule
DFC2020 &  48.6 & \textbf{54.3} & \small \textcolor{green!60!black}{+5.7} \\
Spacenet1 & 75.3  & \textbf{78.2}  &\small \textcolor{green!60!black}{+2.9} \\
Sen1Floods11 & 86.0 & \textbf{89.6} &\small \textcolor{green!60!black}{+3.6}  \\
\midrule
DIOR &  73.7 & \textbf{79.0} & \small \textcolor{green!60!black}{+5.3} \\
\bottomrule
\end{tabular}}
\end{table}

\section{Downstream Datasets Sizes}
Table \ref{tab:datasets} provides an overview of the evaluation datasets used in this work, covering scene classification, segmentation, and object detection tasks. For each dataset, we report the number of classes, label type, input channels, and number of samples. We also include the spatial resolution at which we train our models; when datasets provide images at different native resolutions, we uniformly resize them to the sizes listed in the table to ensure consistent training.

\begin{figure*}
    \centering
    \includegraphics[width=1\linewidth]{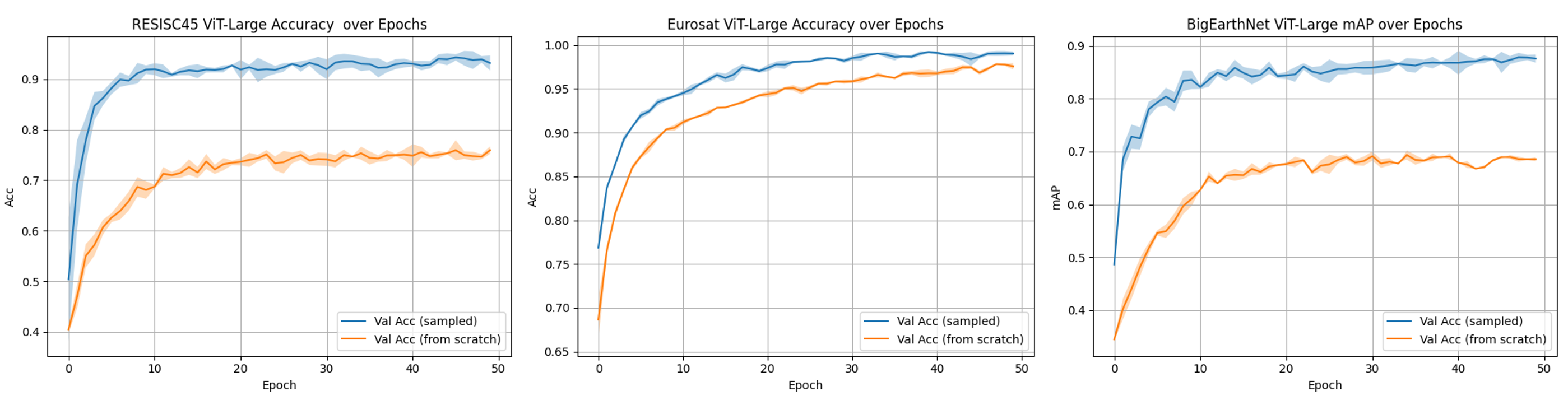}
    \caption{Convergence comparison between GeoSANE-initialized models and models trained from scratch.}
    \label{fig:conv_rates}
\end{figure*}

\begin{table*}[t]
\centering
\small
\begin{tabular}{|l|l|c|c|c|c|c|}
\hline
\textbf{Task} & \textbf{Dataset} & \textbf{\# of classes} & \textbf{Labels (per image)} & \textbf{Size} & \textbf{Channels} & \textbf{\# of samples}\\
\hline

\multirow{6}{*}{\textbf{Classification}} 
& RESISC45~\cite{Cheng2017RemoteSI} & 45  & single  & $256 \times 256$ & RGB & 31.5K \\
& EuroSAT~\cite{Helber2017EuroSATAN}   & 10   & single & $64 \times 64$ & Multispectral& 27K \\
& fMoW~\cite{Christie2017FunctionalMO}  & 63   & single & $512 \times 512$ & Multispectral & $> 1M$\\
& BigEarthNet~\cite{Sumbul2019BigearthnetAL}    & 19  & multiple & $120 \times 120$  & Multispectral & 590K\\
& Sen12Flood~\cite{rambour2020sen12}      & 2 (binary) &  single  & $256 \times 256$ & SAR & 16K\\
& California Wildfires~\cite{california_fire_perimeters}& 2 (binary)  &  single  & $224 \times 224$   & SWIR & 20K\\
\hline

\multirow{3}{*}{\textbf{Segmentation}} 
& DFC2020~\cite{rha7-m332-19}        & 8  &multiple  & $256 \times 256$ & Multispectral & 5K\\
& Sen1Floods11~\cite{Bonafilia2020Sen1Floods11AG}      & 2 (binary)  &    single   & $512 \times 512$   & SAR & 5K\\
& Spacenet1~\cite{Etten2018SpaceNetAR}        & 2 (binary)  & single &  $400 \times 432$  & RGB & 7K\\
\hline

\textbf{Obj. Detection} 
& DIOR~\cite{Li2019ObjectDI}   & 20   & multiple  & $800 \times 800$ & RGB & 23.5K\\
\hline

\end{tabular}
\caption{Overview of datasets, tasks, label types, and channels.}
\label{tab:datasets}
\end{table*}

\begin{figure}
    \centering
    \includegraphics[width=0.85\linewidth]{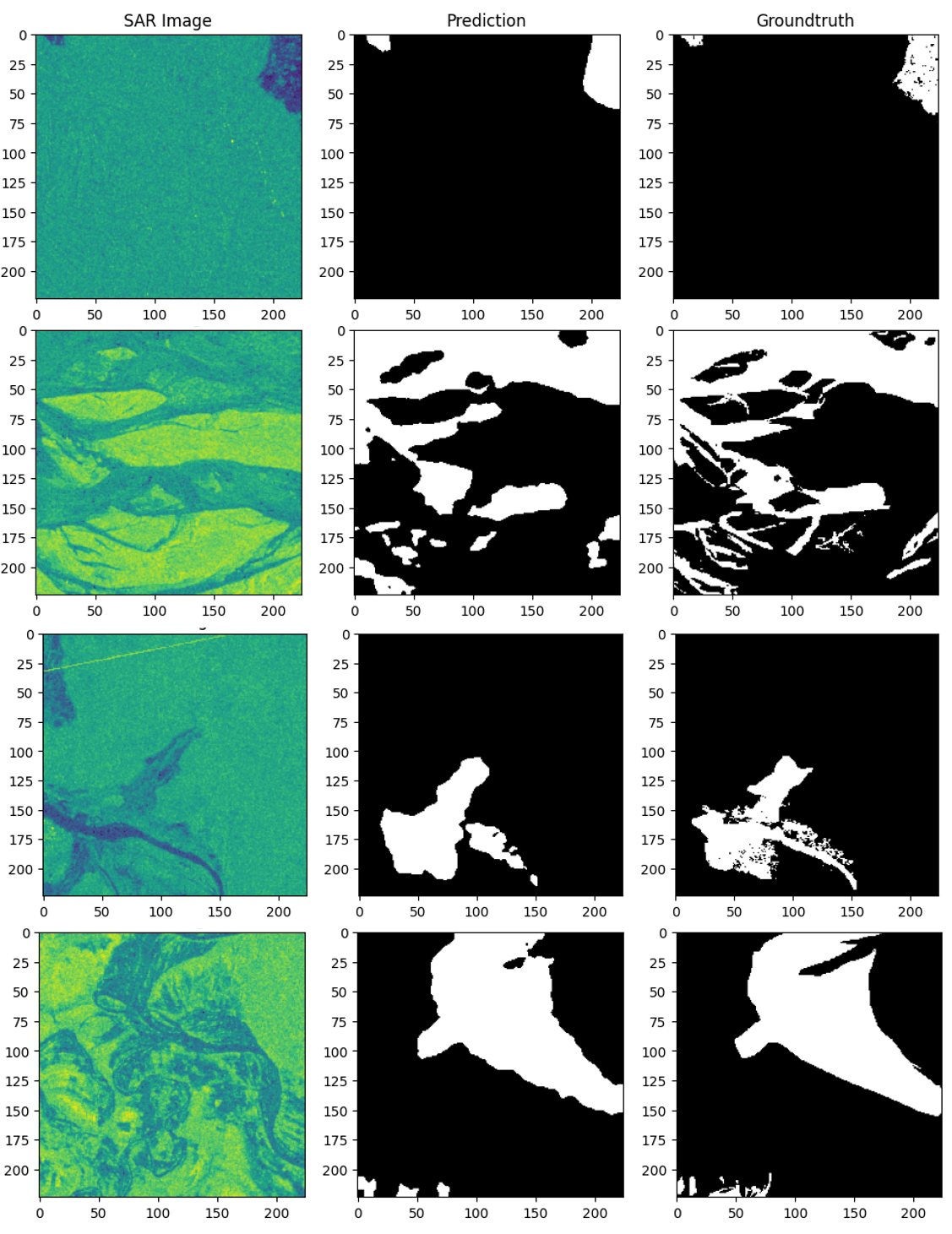}
    \caption{Qualitative Results of the Flood Segmentation task  on the Sen1Floods11 dataset}
    \label{fig:segmentation}
\end{figure}


\section{Additional Details on Model Collection Retrieval and Automatic Loader}
\vspace{-0.25cm}
\begin{lstlisting}[language=bash, caption={Tags and keywords used for models retrieval.}, label={lst:keywords}]
[remote sensing, remote-sensing, NDVI, DSM,
remotesensing, earth observation, enMAP,
earth-observation, EO, satellite,  LiDAR,
satellites, satellite imagery, S1, S2,
satellite-imagery, aerial, aerial imagery, 
aerial-imagery, Sentinel-1, Sentinel 1, 
Sentinel-2, Sentinel 2, SAR, landsat, MODIS,
synthetic-aperture-radar, multispectral, 
multi-spectral, hyperspectral, cloud-mask, 
hyper-spectral, enmap, vegetation-index,  AVIRIS, VIIRS, Pleiades, PlanetScope,
WorldView, drone, UAV, CubeSat, rsfm, RSFM,
satMAE, scalemae, satmae, dofa, optical,
radar, landcover, land-cover, land cover,
land use, land-use, urban mapping,
urban land cover, deforestation, snow cover,
flood detection, flood mapping, wildfire,
fire detection, glacier monitoring, DFC2023,
hazard mapping, coastal erosion, crop 
monitoring, precision agriculture, air
quality, natural disasters, marine debris, 
disaster response, soil sealing, methane
detection, building extraction, road 
extraction, ai4eo, ml4eo, torchgeo, eo-learn,
rasterio, geopandas, earthengine, EuroSAT,
BigEarthNet, xView, FloodNet, SpaceNet,
so2sat, bigearthnet, brick-kiln, forestnet,
pv4ger, RESISC-45, pv4ger-seg, chesapeake,
cashew-plant, Crop Types, NeonTree, Cattles,
SegMunich, UC Merced, AID, FAIR1M, DIOR,
iSAID, ISPRS Potsdam, LEVIR-CD, BurnScars,
MADOS, PASTIS, Sen1Floods11, DynamicEarthNet, 
FiveBillionPixels, CTM-SS, SpaceNet7, NAIP, 
AI4Farms, METER-ML, fMoW, MLRSNet, WHU-RS19, 
Optimal-31, AiRound, CV-BrCT, SpaceNet2, 
INRIA Aerial, GID-15, DFC2020, Dynamic World,
MARIDA, WHU Aerial, Vaihingen, OSCD, DSIFN]
\end{lstlisting}

\begin{figure*}[t]
    \centering
    \includegraphics[width=1.0\linewidth]{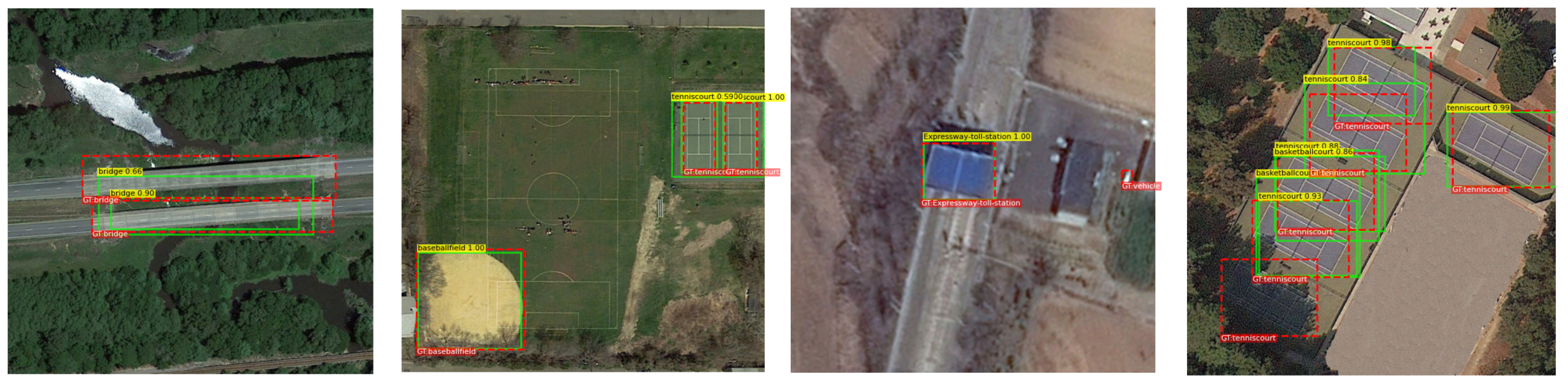}
    \caption{Qualitative Results of the Object Detection task on DIOR dataset}
    \label{fig:detection}
\end{figure*}

As outlined in the main paper, we build the GeoSANE model collection by querying the HuggingFace Hub using a comprehensive set of modality-, task-, and dataset-specific keywords (see Listing \ref{lst:keywords}). Here we provide the additional technical details required for full reproducibility. Because many remote sensing repositories on HuggingFace lack consistent metadata~\cite{horwitz2025we}, we extend the search beyond tags by matching the same keyword vocabulary against repository names, model filenames, and model-card text. This procedure allows us to recover models even when authors provide incomplete or missing tags. The full list of retrieval keywords used in our search is included above.

A second practical challenge is that a substantial fraction of the collected models (particularly those originating from TorchGeo, SSL4EO, or various institutional releases) do not provide configuration files or explicit architectural specifications. To ensure that all such models can be loaded in a unified manner, we implement an automatic architecture reconstruction mechanism. Instead of relying on external configs, the loader infers the backbone type and input dimensionality directly from the structure of the \texttt{state\_dict}. Convolutional backbones (e.g., ResNet-type models) are identified through their stem convolutions, whose tensor shapes reveal both the expected number of input channels and the architectural family. Transformer-based models (e.g., ViT, Swin) are detected through their patch-embedding layers, from which the loader extracts both \texttt{in\_chans} and the embedding dimension. When the checkpoint structure remains ambiguous, we fall back on controlled filename heuristics (e.g., detecting multispectral, SAR, RGB, or mission-specific Sentinel/Landsat naming patterns) to infer sensing modality and band count. Using these inferred attributes, the loader reconstructs the closest matching architecture from \texttt{timm} or \texttt{segmentation\_models\_pytorch}, and loads the checkpoint. This procedure enables GeoSANE to handle heterogeneous, partially documented checkpoints in a consistent and fully automated way. All loader code will be released alongside the final model collection.


\section{Qualitative Results}
Aside from quantitative results, we also provide qualitative examples. Following the same procedure described in the experimental setup, we first generate a Swin-B backbone with GeoSANE, then attach the appropriate task-specific head (a Faster R-CNN detection head for DIOR, or a segmentation head for Sen1Floods11). 
Figure~\ref{fig:detection} shows some object detection outputs on the DIOR dataset, while Figure~\ref{fig:segmentation} displays semantic flood segmentation results on Sen1Floods11. 

\section{UMAP Visualization of the Latent Weight Space}
\label{sec:supp:umap}
To better understand the structures learned by GeoSANE, we visualize (Figure~\ref{fig:umaps_horizontal}) the latent representations of all models in our remote sensing collection using UMAP. For each model, we extract its full latent embedding sequence, sample 100 tokens randomly, and project these vectors to 2D using UMAP. When colored by architecture (top), the embedding shows clear structural organization: models sharing similar backbone types, such as ViTs, UNets, Swins or ResNets form well-separated clusters. This indicates that GeoSANE’s latent space preserves meaningful distinctions between architectural families. When colored by sensing modality (bottom), the same embedding shows modality-related structure: models having similar inputs (e.g., SAR or Landsat ) tend to appear in nearby regions of the space. 
Together, these observations show that GeoSANE organizes heterogeneous remote sensing models into meaningful latent groups.

\section{Convergence Behavior of GeoSANE-generated Models}
We study the convergence behavior of models initialized with GeoSANE compared to models trained from scratch. Figure~\ref{fig:conv_rates} reports the validation performance over epochs for three representative datasets (RESISC45, EuroSAT, BigEarthNet), all using a ViT-L backbone. In every case, GeoSANE-generated initializations achieve strong accuracy within the first few epochs and maintain a consistent lead throughout training. This shows that GeoSANE does not only produce models competitive with remote sensing foundation models, but it also consistently accelerates optimization; which is an important advantage in settings with limited compute or training budgets.


\section{Model Checkpoints Used for Training}
\label{app:model_list}

Table~\ref{tab:model_list} provides the complete list of model checkpoints used to train GeoSANE. For each model, we report the primary task (Table~\ref{tab:tasks}) and the sensing modalities (Table~\ref{tab:modalities}) used during training when this information is available from the original repository or documentation.

\begin{table*}
\centering
\scriptsize
\setlength{\tabcolsep}{3pt}
\scalebox{0.8}{
\begin{tabular}{lll}
\toprule
Checkpoint & Checkpoint & Checkpoint \\
\midrule
\texttt{\detokenize{jaychempan__EarthSynth}} & \texttt{\detokenize{azdin__llava-onevision-weather-dora}} & \texttt{\detokenize{mayrajeo__marine-vessel-yolo}} \\
\texttt{\detokenize{torchgeo__presto}} & \texttt{\detokenize{azdin__llava-onevision-weather-qlora}} & \texttt{\detokenize{torchgeo__core-dino}} \\
\texttt{\detokenize{DevPanda004__PrithviFlood}} & \texttt{\detokenize{azdin__qwen2-vl-weather-adalora}} & \texttt{\detokenize{torchgeo__delineate-anything}} \\
\texttt{\detokenize{Mahadih534__YoloV8-VisDrone}} & \texttt{\detokenize{azdin__qwen2-vl-weather-dora}} & \texttt{\detokenize{torchgeo__delineate-anything-s}} \\
\texttt{\detokenize{Mahadih534__yolov8_ship_det_satellite}} & \texttt{\detokenize{banghyunmin__Thermal_Video_Detection}} & \texttt{\detokenize{torchgeo__yolo11s_marine_vessel_detection}} \\
\texttt{\detokenize{RedbeardNZ__LatentSync-1.6}} & \texttt{\detokenize{banghyunmin__thermal-people-yolov11n}} & \texttt{\detokenize{wangyi111__Copernicus-FM}} \\
\texttt{\detokenize{azdin__llava-onevision-weather-adalora}} & \texttt{\detokenize{mayrajeo__marine-vessel-detection-yolov8}} & \texttt{\detokenize{IGNF__FLAIR-HUB_LC-A_IR_convnextv2base-unet}} \\
\texttt{\detokenize{IGNF__FLAIR-HUB_LC-A_IR_convnextv2base-upernet}} & \texttt{\detokenize{IGNF__FLAIR-HUB_LC-A_IR_convnextv2tiny-upernet}} & \texttt{\detokenize{IGNF__FLAIR-HUB_LC-A_IR_swinbase-unet}} \\
\texttt{\detokenize{IGNF__FLAIR-HUB_LC-A_IR_swinbase-upernet}} & \texttt{\detokenize{IGNF__FLAIR-HUB_LC-A_IR_swinlarge-upernet}} & \texttt{\detokenize{IGNF__FLAIR-HUB_LC-A_IR_swinsmall-upernet}} \\
\texttt{\detokenize{IGNF__FLAIR-HUB_LC-A_IR_swintiny-upernet}} & \texttt{\detokenize{IGNF__FLAIR-HUB_LC-A_RGB_swinbase-upernet}} & \texttt{\detokenize{IGNF__FLAIR-HUB_LC-A_RGB_swinlarge-upernet}} \\
\texttt{\detokenize{IGNF__FLAIR-HUB_LC-A_RGB_swinsmall-upernet}} & \texttt{\detokenize{IGNF__FLAIR-HUB_LC-A_RGB_swintiny-upernet}} & \texttt{\detokenize{IGNF__FLAIR-HUB_LC-D_swinbase-upernet}} \\
\texttt{\detokenize{IGNF__FLAIR-HUB_LC-F_swinbase-upernet}} & \texttt{\detokenize{IGNF__FLAIR-HUB_LC-I_swinbase-upernet}} & \texttt{\detokenize{IGNF__FLAIR-HUB_LC-L_swinbase-upernet}} \\
\texttt{\detokenize{IGNF__FLAIR-HUB_LPIS-A_swinbase-upernet}} & \texttt{\detokenize{IGNF__FLAIR-HUB_LPIS-I_swinbase-upernet}} & \texttt{\detokenize{IGNF__FLAIR-HUB_LPIS-J_swinbase-upernet}} \\
\texttt{\detokenize{Jabasingh__VCTI}} & \texttt{\detokenize{torchgeo__core-dino}} & \texttt{\detokenize{torchgeo__satlas}} \\
\texttt{\detokenize{chrimerss__flood-foundation-prithvi-100m}} & \texttt{\detokenize{torchgeo__croma}} & \texttt{\detokenize{torchgeo__seco-eco}} \\
\texttt{\detokenize{chrimerss__flood-foundation-prithvi-300m}} & \texttt{\detokenize{torchgeo__decur}} & \texttt{\detokenize{torchgeo__seco-eco-ndvi}} \\
\texttt{\detokenize{chrimerss__flood-foundation-prithvi-600m}} & \texttt{\detokenize{torchgeo__delineate-anything}} & \texttt{\detokenize{torchgeo__sentinel1_unet_effb4_openearthmap_sar}} \\
\texttt{\detokenize{chrimerss__flood-foundation-prithvi-tiny}} & \texttt{\detokenize{torchgeo__delineate-anything-s}} & \texttt{\detokenize{torchgeo__ssl4eo_landsat}} \\
\texttt{\detokenize{chrimerss__flood-foundation-resnet101-unet}} & \texttt{\detokenize{torchgeo__dofa}} & \texttt{\detokenize{torchgeo__swin_v2_b_naip_rgb_satlas}} \\
\texttt{\detokenize{chrimerss__flood-foundation-resnet152-unet}} & \texttt{\detokenize{torchgeo__earthloc}} & \texttt{\detokenize{torchgeo__swin_v2_b_sentinel2_rgb_satlas}} \\
\texttt{\detokenize{chrimerss__flood-foundation-resnet50-unet}} & \texttt{\detokenize{torchgeo__fields-of-the-world}} & \texttt{\detokenize{torchgeo__unet_resnet34_oam_rgb_tcd}} \\
\texttt{\detokenize{galeio-research__OceanSAR-1}} & \texttt{\detokenize{torchgeo__ftw}} & \texttt{\detokenize{torchgeo__unet_resnet50_oam_rgb_tcd}} \\
\texttt{\detokenize{galeio-research__OceanSAR-1-tengeop}} & \texttt{\detokenize{torchgeo__resnet18_sentinel2_all_moco}} & \texttt{\detokenize{torchgeo__vit_base_patch32_224_skyclip_50pct}} \\
\texttt{\detokenize{galeio-research__OceanSAR-1-wave}} & \texttt{\detokenize{torchgeo__resnet18_sentinel2_rgb_moco}} & \texttt{\detokenize{torchgeo__vit_large_patch14_224_clip_laionrs}} \\
\texttt{\detokenize{galeio-research__OceanSAR-1-wind}} & \texttt{\detokenize{torchgeo__resnet18_sentinel2_rgb_seco}} & \texttt{\detokenize{torchgeo__vit_large_patch14_224_skyclip_30pct}} \\
\texttt{\detokenize{mrm8488__convnext-tiny-finetuned-eurosat}} & \texttt{\detokenize{torchgeo__resnet50_fmow_rgb_gassl}} & \texttt{\detokenize{torchgeo__vit_large_patch14_224_skyclip_50pct}} \\
\texttt{\detokenize{openclimatefix__power_perceiver}} & \texttt{\detokenize{torchgeo__resnet50_landsat7_l2_all_moco}} & \texttt{\detokenize{torchgeo__vit_large_patch16_224_fmow_rgb_scalemae}} \\
\texttt{\detokenize{pszemraj__convnextv2-nano-22k-384-boulderspot}} & \texttt{\detokenize{torchgeo__resnet50_sentinel1_all_moco}} & \texttt{\detokenize{torchgeo__vit_small_patch16_224_sentinel2_all_dino}} \\
\texttt{\detokenize{quantum-leap-vcti__VCTI-RoBERTa-Fiber}} & \texttt{\detokenize{torchgeo__resnet50_sentinel2_all_dino}} & \texttt{\detokenize{torchgeo__vit_small_patch16_224_sentinel2_all_moco}} \\
\texttt{\detokenize{swardiantara__drone-term-extractor}} & \texttt{\detokenize{torchgeo__resnet50_sentinel2_all_moco}} & \texttt{\detokenize{torchgeo__yolo11s_marine_vessel_detection}} \\
\texttt{\detokenize{torchgeo__ai4g_flood}} & \texttt{\detokenize{torchgeo__resnet50_sentinel2_rgb_moco}} & \texttt{\detokenize{Burdenthrive__cloud-detection-segformer-mit_b4-RGB}} \\
\texttt{\detokenize{torchgeo__copernicus-fm}} & \texttt{\detokenize{torchgeo__resnet50_sentinel2_rgb_seco}} & \texttt{\detokenize{Burdenthrive__cloud-detection-unet-regnetzd8}}\\
\texttt{\detokenize{ingmarnitze__thaw-slump-segmentation}} & \texttt{\detokenize{isaaccorley__unet_resnet50_oam_rgb_tcd}}& \texttt{\detokenize{Kaludi__CSGO-Minimap-Layout-Generation}}\\
\texttt{\detokenize{gsumbul__SMARTIES-v1-finetuned-models}} & \texttt{\detokenize{truthdotphd__cloud-detection}}& \texttt{\detokenize{DarthReca__actu-magnitude-regression}}\\
\texttt{\detokenize{DimitrisMantas__RoofSense}}& & \\
\bottomrule
\end{tabular}}
\caption{Remote Sensing Model checkpoints used to train GeoSANE.}
\label{tab:model_list}
\end{table*}

\begin{table}[h]
\centering
\small
\begin{tabular}{lc}
\toprule
Modality & Number of Models \\
\midrule
Multispectral & 41 \\
RGB & 15 \\
Multimodal & 13 \\
SAR & 6 \\
DEM & 2 \\
Unknown & 26 \\
\bottomrule
\end{tabular}
\caption{Distribution of modalities for the models used to train GeoSANE.}
\label{tab:modalities}
\end{table}

\begin{table}[h]
\centering
\small
\begin{tabular}{lc}
\toprule
Task & Number of Models \\
\midrule
Self-supervised representation learning & 36 \\
Semantic segmentation & 19 \\
Object detection & 7 \\
Classification & 4 \\
Generative models & 3 \\
Regression & 2 \\
Unknown & 32 \\
\bottomrule
\end{tabular}
\caption{Distribution of primary tasks for the models used to train GeoSANE.}
\label{tab:tasks}
\end{table}

\end{document}